\documentclass{article}

\usepackage{arxiv}
\usepackage[utf8]{inputenc}
\usepackage[T1]{fontenc}
\usepackage{mathtools}
\usepackage{amsthm}
\usepackage{amssymb}
\usepackage[numbers]{natbib}
\usepackage[dvipsnames]{xcolor}
\usepackage[colorlinks=true, linktoc=page, linkcolor={blue}, citecolor={blue}, urlcolor={cyan}]{hyperref}
\usepackage{cleveref}
\usepackage{graphicx}
\usepackage{enumitem}
\usepackage{booktabs}
\usepackage{caption}
\usepackage{subcaption}
\usepackage{multirow}
\usepackage[ruled,vlined]{algorithm2e}
\usepackage{adjustbox}

\graphicspath{{figs}}
\newtheorem{assumption}{Assumption}

\title{Context Representation via Action-Free Transformer
encoder-decoder for Meta Reinforcement Learning}

\author{Amir Mehdi Soufi Enayati \\
Department of Mechanical Engineering \\
University of Victoria \\
3800 Finnerty Road \\
Victoria, BC, Canada \\
\texttt{amsoufi@uvic.ca} \\
\And
Homayoun Honari \\
Mila-Quebec AI Institute \\
6666, St-Urbain \\
Montreal, QC, Canada \\
\texttt{homayoun.honari@mila.quebec} \\
\And
Homayoun Najjaran$^*$ \\
Department of Mechanical Engineering \\
University of Victoria \\
3800 Finnerty Road \\
Victoria, BC, Canada \\
\texttt{najjaran@uvic.ca} \\
}

\date{}

\renewcommand{\headeright}{}
\renewcommand{\undertitle}{}
\renewcommand{\shorttitle}{Context Representation via Action-Free Transformer encoder-decoder for Meta-RL}

\begin{document}
\maketitle

\begin{abstract}
Reinforcement learning (RL) enables robots to operate in uncertain environments, but standard approaches often struggle with poor generalization to unseen tasks. Context-adaptive meta reinforcement learning addresses these limitations by conditioning on the task representation, yet they mostly rely on complete action information in the experience making task inference tightly coupled to a specific policy. This paper introduces \textbf{C}ontext \textbf{R}epresentation via \textbf{A}ction-\textbf{F}ree \textbf{T}ransformer encoder–decoder (\textbf{CRAFT}), a belief model that infers task representations solely from sequences of states and rewards. By removing the dependence on actions, CRAFT decouples task inference from policy optimization, supports modular training, and leverages amortized variational inference for scalable belief updates. Built on a transformer encoder–decoder with rotary positional embeddings, the model captures long-range temporal dependencies and robustly encodes both parametric and non-parametric task variations. Experiments on the MetaWorld ML-10 robotic manipulation benchmark show that CRAFT achieves faster adaptation, improved generalization, and more effective exploration compared to context-adaptive meta-RL baselines. These findings highlight the potential of action-free inference as a foundation for scalable RL in robotic control.\footnote{This work has been published in part in the doctoral dissertation of Amir M. Soufi Enayati~\cite{enayati2025practical}.}
\end{abstract}


\section{Introduction}\label{sec:meta_intro}
Reinforcement learning (RL) is traditionally framed as the problem of finding an optimal policy that maximizes expected return in a Markov decision process (MDP). However, real-world applications introduce complexities beyond this formulation, especially in unknown environments where balancing exploration and exploitation becomes critical. This challenge is magnified in robotics as data collection is costly, and sample efficiency is paramount.

Biological systems demonstrate adaptability by generalizing experience across tasks, i.e., a capability also desirable in robotic agents. While standard RL assumes known rewards and transitions, practical applications involve dynamic or unknown structures. Meta-reinforcement learning (meta-RL) addresses this by enabling agents to ``\textit{learn how to learn}'' capturing task structures for better generalization.

\begin{figure}[ht]
    \centering
    \includegraphics[width=\columnwidth]{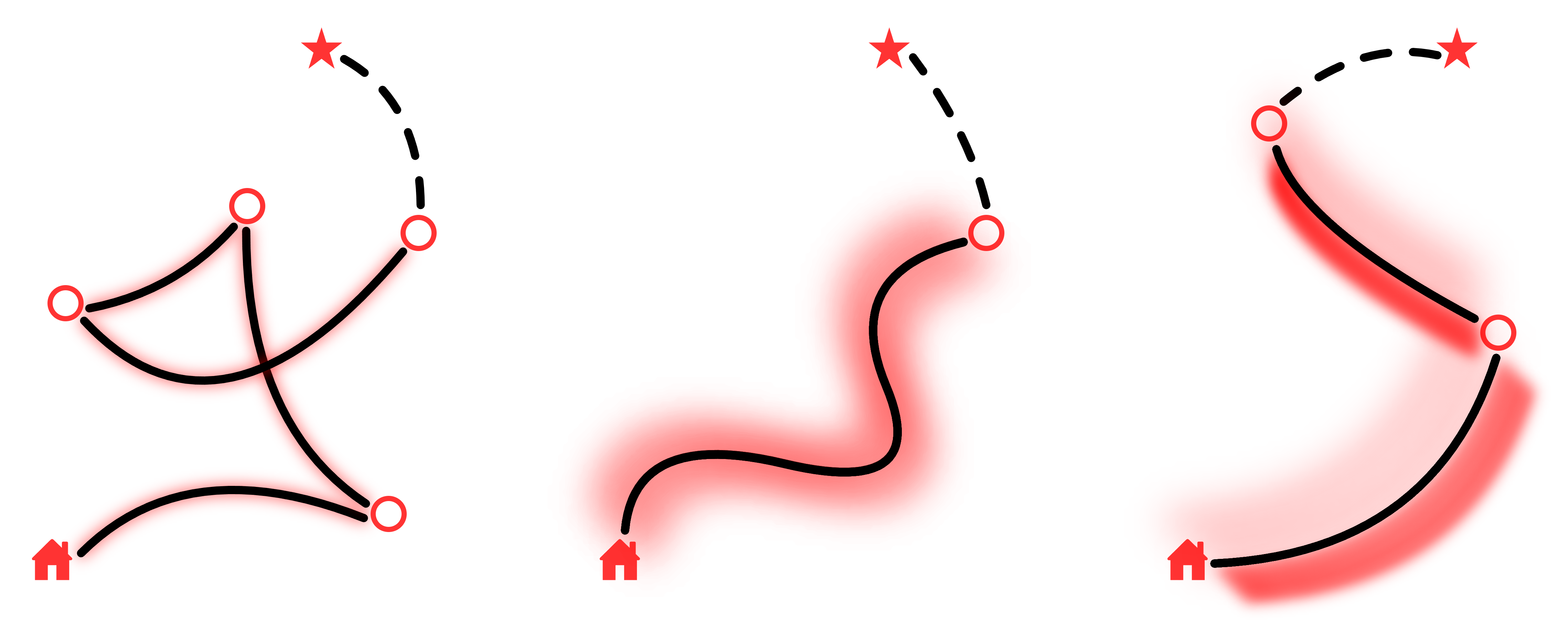}
    \caption{Comparison of belief construction in meta-RL. Posterior sampling (left) infers a latent belief from trajectories to compress relevant information. Bayes-optimal agents (middle) update the full posterior over tasks using complete interaction histories, which is intractable in complex domains. Bayes-adaptive formulations (right) embed belief into the state with principled belief updates; the proposed method follows this formulation.}
    \label{fig:meta-bamdp}
\end{figure}

Although Bayes-optimal agents offer ideal exploration-exploitation trade-offs, they are intractable in complex domains. Posterior sampling and latent variable meta-learning attempt to approximate this behavior, encoding task uncertainty into structured representations to reduce reliance on extensive interactions. Yet, most meta-RL approaches struggle with significant task variability, often assuming smooth parametric changes between tasks, which limits their adaptability in real-world robotic scenarios involving non-parametric variations \cite{zintgraf2019varibad}. A schematic of different approaches in context accumulation for exploration in meta-RL is depicted in \Cref{fig:meta-bamdp}.

Algorithms like the Model-Agnostic Meta-Learning (MAML)~\cite{finn2017model} and RL$^2$ \cite{duan2016rl2} focus on fast adaptation, but they often falter in handling long-term dependencies and generalization across unseen tasks. Recent advances in transformer architectures have shown promise in this context. Transformers, originally introduced for Natural Language Processing (NLP), excel in sequence modeling, offering superior capabilities over RNNs in capturing long-term dependencies. As an example, TrMRL \cite{melo2022transformers} leverages transformer architecture to derive an intraepisodic understanding of the environment to enhance generalization but disregards the variational belief inference in conventional Bayes-adaptive meta-RL \cite{zintgraf2019varibad} that provides meaningful feature extraction on the task distribution.

The attention mechanism in Transformers allows for efficient encoding of sequential experiences without the vanishing gradient issues of RNNs. This ability facilitates modeling of meta-features within or along with the policy and will improve adaptation and robustness. Moreover, transformers remove the need for handcrafted structures, facilitating scalable and flexible architecture in high-dimensional, dynamically changing environments. TrMRL extends this by modeling memory reinstatement via associating multi-layer working memories. This structure attempts to minimize Bayes' risk by reducing the condition to only the most recent interactions.

To address current limitations, a framework is proposed in this paper for extracting long-term interepisodic features for Bayes-adaptive exploration. Additionally, CRAFT isolates the agent's performance from task history by removing the condition of the belief model on the actions. This abstraction will allow learned behaviors to be reused and recombined in every stage of the training. Leveraging structured latent spaces and variational inference, the method facilitates strategic exploration and robust generalization, evaluated in a robotic simulation benchmark. This integration of probabilistic task embeddings and meta-learning techniques aims to ameliorate performance limitations and advance autonomy in RL-driven robot manipulation.

The remainder of this paper is structured as follows: \Cref{sec:meta_background} provides foundational background. \Cref{sec:meta_method} details the CRAFT method, including the inference mechanisms and decomposition strategy. \Cref{sec:meta_related} positions the work within the literature. \Cref{sec:meta_finding} presents experiments. Finally, \Cref{sec:meta_conclusion} concludes, and \Cref{sec:meta_future} outlines future directions.

\section{Background}\label{sec:meta_background}

\subsection{Meta-RL General Outline}
Meta-RL is concerned with finding a policy, after seeing $n \in N$ tasks (each denoted by $\mathcal{T}_{\mathrm{i}}$, $1 \leq i \leq N$) and their corresponding environments during training, that is optimal in the sense that it is close enough to the optimal single-task policy for all of the tasks in training or test. The proximity here is not always intended to be an averaging or trade-off in performance; rather, it can be measured in learning distance, i.e., the number of shots or experiences that policy requires to adapt to a new task. Therefore, a meta-policy is presumably adaptable with a few-shot training to achieve optimal performance in each $\mathcal{T}_{\mathrm{i}}$.
In general, these tasks can be different stages of one single time-variant environment, multiple parameterically different time-invariant environments, or any two different environments sharing some fundamental dynamic or semantic features \cite{beck2023survey}.


MAML is a conventional baseline that many other implementations are based on. It is a gradient-based optimization problem to formulate proximity after a few adaptation steps. With the loss function for each task $\mathcal{T}_{\mathrm{i}}$ defined as
\begin{equation}
\mathcal{L}_{\mathcal{T}_{\mathrm{i}}}(\pi_{\theta}) = -\mathbb{E}_{(s_t,a_t) \sim \pi_{\theta}} \left[\sum_{t=0}^{N_{\tau}}\gamma^t r_{t+1}\right],
\end{equation}
a new parameter defining interim policy improvement for $\mathcal{T}_i$ can be calculated in the meta-training stage with
\begin{equation}
\theta'_i = \theta - \alpha \nabla_{\theta} \mathcal{L}_{\mathcal{T}_{\mathrm{i}}}(\pi_{\theta}).
\end{equation}

The resulting interim policy $\pi_{\theta'_i}$ will then be used to collect new trajectory experiences and losses for updating the policy in the meta-training stage through
\begin{equation}
\theta = \theta - \beta \nabla_{\theta} \sum_{\mathcal{T}_{\mathrm{i}}} \mathcal{L}_{\mathcal{T}_{\mathrm{i}}}(\pi_{\theta'_i}).
\end{equation}

Both $\alpha$ and $\beta$ are learning rate hyperparameters of MAML. By excluding other tasks from the task distribution or adding a new task, $\theta^*_i$ or $\theta^*$, in general, can be trained after a few further iterations of meta-training. $\pi_{\theta^*}$ is deemed the optimal policy for task $\mathcal{T}$.

Meta-RL faces several challenges that hinder its adoption and effectiveness. One major challenge is the issue of sample efficiency. Meta-RL algorithms typically require a large number of interactions with the environment to learn useful meta-policies. However, collecting such a vast amount of data can be time-consuming and resource-intensive, especially in complex and high-dimensional environments~\cite{rakelly2019efficient}.

Another challenge in meta-RL is the problem of generalization across tasks and domains. Meta-RL algorithms aim to learn meta-policies that can adapt to new tasks or environments with minimal additional training. However, achieving robust generalization remains a significant issue. Meta-learned policies often struggle to transfer knowledge to tasks or domains that differ significantly from the training distribution, resulting in poor performance and limited applicability. Yu et al.~\cite{yu2020meta} have shown that state-of-the-art meta-RL algorithms usually struggle in adaptation to an unseen robotic manipulation task after being exposed to ten or even 45 tasks in meta-training. This indicates that knowledge sharing among the training tasks is not easily generalizable to the test domain.

Addressing these challenges requires developing effective methods to capture variations. Incorporating domain adaptation techniques and exploring approaches that can extract structured knowledge from an unstructured training might enhance the generalization capability.

\subsection{Bayesian Approach To Context-Adaptive Meta-RL}
Meta-RL aims to develop agents capable of rapid adaptation to new tasks by leveraging prior experiences across a distribution of related Markov Decision Processes. A key challenge in meta-RL is efficiently balancing exploration and exploitation when the underlying task dynamics are initially unknown. This challenge is even more important when dealing with on-policy algorithms that rely on the most recent interactions with a task and are prone to achieving sub-optimal performance because of the local minima issues or forgetting in the long run. Bayesian meta-reinforcement learning (Bayesian meta-RL) provides a principled framework to address this challenge by maintaining and updating a belief distribution over the latent MDP structure by defining a recent online history, thereby enabling approximate Bayesian optimal behavior. This design also solves the intractable nature of posterior sampling present in the off-policy algorithms.

A prominent model-free meta-RL algorithm, RL$^2$, relies on recurrent networks that implicitly encode task information through sequential interactions \cite{duan2016rl2}. Memory-based methods such as RL$^2$ process past states, rewards, and actions as auxiliary inputs, allowing the agent to adapt its policy within a task dynamically. This recurrent structure aligns closely with Bayesian formulations, where past experiences inform a belief distribution over task parameters \cite{ortega2019meta}. The Bayesian perspective reframes meta-learning as an inference problem, where the agent maintains a posterior belief over task dynamics and updates it continuously based on observed transitions and rewards.

Variational Bayes-Adaptive Deep RL (VariBAD) \cite{zintgraf2019varibad} extends this concept by integrating variational inference into model task uncertainty explicitly. The agent learns a stochastic latent representation of the task, conditioned on past recent experiences, which serves as an inductive bias for exploration-exploitation trade-offs. Unlike memory-based approaches that implicitly infer task dynamics through recurrence, VariBAD encodes the belief in a structured latent space, enabling a more explicit and interpretable task representation.

For a regular stochastic episodic MDP corresponding to the task $\mathcal{T}_i$,
\begin{equation}\label{eq:meta-mdp}
    \mathcal{M} = (\mathcal{S}, \mathcal{A}, \mathcal{P}_R,  \mathcal{P}_S, \mathcal{P}_{S_0}, \gamma, H)
\end{equation}
with state space $\mathcal{S}$, action space $\mathcal{A}$, discount factor $\gamma$, and maximum episode length of $H$, $\mathcal{P}_S^i$ and $\mathcal{P}_R^i$ represent conditional probabilistic distributions $p(s'|s,a)$ and $p(r|s,a)$ respectively defined on the state domain $\mathcal{S}$ and the reward domain $\mathcal{R}$. The successive state and the instant reward are recognized as sampling from $s' \sim p(s'|s,a)$ and $r \sim p(r|s,a)$. Similarly, the Bayes-Adaptive MDP (BAMDP) framework formalizes optimal decision-making under uncertainty by augmenting the state space with a belief distribution over task dynamics \cite{duff2002optimal, ghavamzadeh2015bayesian}. In this setting, the agent’s objective is to maximize expected return while simultaneously refining its belief over the task. Formally, the belief update follows Bayes’ rule, integrating observed transitions into the posterior distribution over the reward and transition functions of the $\mathcal{M}^i$, the MDP representing the task $\mathcal{T}_i$, as
\begin{equation}\label{eq:belief}
    b_t(\mathcal{P}_R^i, \mathcal{P}_S^i) = p(\mathcal{P}_R^i, \mathcal{P}_S^i | \tau_{0:t})
\end{equation}
with \(\tau_{0:t} = \{s_0, a_1, r_1, s_1, \cdots, s_{t-1}, a_t, r_t, s_t\}\) denotes the agent’s trajectory up to timestep $t$. The belief-augmented BAMDP enables structured exploration by guiding the agent toward actions that optimally reduce uncertainty while maximizing reward. However, exact Bayesian inference is intractable in complex environments, necessitating approximate solutions.

By integrating the belief in the state definition, a hyper-state space $\mathcal{S}^+=\mathcal{S}\times\mathcal{B}$ is obtained, where $\mathcal{B}$ is the belief space, and the hyper-state at timestep $t$ is therefore \(s^+_t=(s_t, b_t)\in\mathcal{S}^+\). The resulting dynamic transition hyper-functions $\mathcal{P}_S^+$ and $\mathcal{P}_R^+$ is derived as
\begin{align}
	& \mathcal{P}_S^+(s^+_{t+1} | s_t^+, a_{t+1}, r_{t+1}) \nonumber \\
	&= \mathcal{P}_S^+(s_{t+1}, b_{t+1}| s_t, a_{t+1}, r_{t+1}, b_t) \nonumber \\
	&= \mathcal{P}_S^+(s_{t+1} | s_t, a_{t+1}, b_t) ~  p(b_{t+1} | s_t, a_{t+1}, r_{t+1}, s_{t+1}, b_t) \nonumber \\
	&= \mathbb{E}_{b_t}\left[ \mathcal{P}_S(s_{t+1} | s_t, a_{t+1}) \right] ~ \delta \Bigl( b_{t+1}=p(\mathcal{P}_R, \mathcal{P}_S|\tau_{0:t+1})\Bigl).
	\label{eq:bamdp_transition}
\end{align}

The consequent state of the original MDP $s_t$ is found via deterministic Bayes' rule applied to the current posterior distribution of the transition function and the belief distribution. Similarly, the reward hyper-function is the expected reward under the current posterior, after the state transition, over reward functions
\begin{align}
\mathcal{P}_R^+(r_{t+1}|s_t^+, a_t, s^+_{t+1})
&= \mathcal{P}_R^+(r_{t+1}|s_t, b_t, a_t, s_{t+1}, b_{t+1})  \nonumber \\
&= \mathbb{E}_{b_{t+1}} \left[ \mathcal{P}_R(r_{t+1}|s_t, a_t, s_{t+1}) \right].
\label{eq:bamdp_reward}
\end{align}
These two dynamic hyper-functions result in defining a BAMDP \cite{duff2002optimal}:
\begin{equation*}
\mathcal{M}^+ = (\mathcal{S^+}, \mathcal{A}, \mathcal{P}_R^+,  \mathcal{P}_S^+, \mathcal{P}_{S_0}^+, \gamma, H^+),
\end{equation*}

\noindent which is a special case of a context-adaptive MDP formed by augmenting Markov states by the posterior beliefs in a partially observable MDP \cite{cassandra1996acting}. The optimization objective of the agent is now to maximize the following for BAMDP
\begin{equation} \label{eq:bamdp_objective}
	\mathcal{J}^+(\pi) = 
	\mathbb{E}_{b_0, \mathcal{P}_{S_0}^+, \mathcal{P}_S^+, \pi} \left[ \sum_{t=0}^{H^+-1} \gamma^t \mathcal{P}_R^+(r_{t+1} | s^+_t, a_t, s^+_{t+1}) \right],
\end{equation}
where the scope of the learning objective is extended over the meta-horizon $H^+$, distinct from the original MDP horizon $H$. These two parameters can naturally coincide; however, longer memory for BAMDP helps maintain the belief under less uncertainty. In other words, remembering a \textit{few} more shots, a few more experiences in the case of RL, will narrow down the belief distribution, so it is reasonable to assume $H^+=n_H \times H$. In this formulation, the exploration-exploitation switch is triggered by the agent when it deems information-seeking actions to be sufficient.

VariBAD \cite{zintgraf2019varibad} applies variational inference to estimate the belief distribution over tasks. Rather than using an explicit belief update, they use an amortized inference network conditioned on previous transitions to generate a variational posterior latent vector. This gives a more flexible online adaptation capability to VariBAD, and as a result, a greater sample efficiency compared to conventional gradient-based meta-RL methods like MAML.

\subsection{Transformer Models}
Transformers \cite{vaswani2017attention} offer a sequence modeling framework based on self-attention, allowing for efficient parallel processing of tokens and long-range dependency capture. The core mechanism employs learned projections into query (\(\mathbf{q}\)), key (\(\mathbf{k}\)), and value (\(\mathbf{v}\)) vectors, with scaled dot-product attention
\begin{equation}
    \text{Attn}(\mathbf{q}, \mathbf{k}, \mathbf{v}) = \text{softmax}\left(\frac{\mathbf{q} \mathbf{k}^\top}{\sqrt{d_k}}\right) \mathbf{v}.
\end{equation}
Encoders apply self-attention across the sequence; decoders use causal masks. This primer supports the rotary positional embedding details that follow in \Cref{sec:meta_method}.

\section{Design and Architecture of CRAFT}\label{sec:meta_method}
This section describes how CRAFT is learning a belief model in context-adaptive meta-RL. The design leverages a transformer encoder-decoder to maintain a posterior belief about the task from action-free trajectories, isolates inference from policy optimization, and supports Bayesian task inference via amortized variational learning.

\subsection{Learning the Belief Representation}\label{sec:meta_encdec}
Whenever, in context-adaptive meta-reinforcement learning, the task inference module is explicitly separated from the policy, such as in TIGR \cite{bing2023tigr} or VariBAD, additional loss functions are typically required to promote meaningful representation learning within the inference model. This is because, in the absence of policy supervision signals, the task representation must be learned from auxiliary objectives, such as reconstructing rewards, transitions, or other relevant statistics of the task environment.

\subsubsection{Task Decomposition}
The representation learned by the inference model, often a latent belief or embedding, serves as a compact and informative summary of the current task. This representation should not only differentiate between tasks but ideally capture internal task structure in a form that is decomposable either temporally or structurally. In meta-RL, such decomposability is critical for generalization and transfer. These subcomponents of tasks are commonly referred to as \textit{skills}, temporally extended, goal-directed policies that serve as reusable building blocks across multiple tasks \cite{hausman2018learning}. Formally, each task \( \mathcal{T}_i \) can be expressed as a sequence of skills
\begin{equation}
    \mathcal{T}_i = \{ \kappa_j \}_{j=1}^{N^i_\kappa}, \quad \kappa_j : \mathcal{S} \times \Theta_\kappa \rightarrow \mathcal{U},
\end{equation}
where \( \mathcal{S} \) denotes the state space, \( \mathcal{U} \) the action space, and \( \Theta_\kappa \) the parameter space of the skill policies. Skills \( \kappa_j \) may be shared across different tasks, enabling modular policy construction and improved sample efficiency in adaptation.

Task representation learning can be supervised using either privileged information or unsupervised reconstruction-based signals. For example, TIGR uses explicit task labels as ground truth supervision. While such privileged supervision can enforce structure, it is often unavailable or too rigid in practical applications. An alternative, more general approach is to learn the task embedding by reconstructing observed transitions and rewards. Although less constrained, this strategy encourages the model to reflect the underlying dynamics and optimality characteristics of the task, which are more aligned with real-world scenarios where task labels are absent.

\subsubsection{Gradient Isolation}
A defining feature of task-inference-based meta-RL algorithms like VariBAD, unlike approaches such as PEARL \cite{rakelly2019efficient}, is the deliberate prevention of policy gradients from backpropagating through the task inference model. This design choice is motivated by several important considerations:

\begin{enumerate}[label=\Roman*.]
    \item \textit{Stable and Modular Belief Inference:} if policy gradients are allowed to flow into the inference model, the resulting latent embeddings may become entangled with the current policy. This coupling leads to task representations that are biased toward short-term policy performance rather than general task structure. By isolating gradient flow, the inference model remains a standalone module that encodes consistent and reusable task representations, independent of the policy’s current behavior.

    \item \textit{Amortized Variational Inference Stability:} in VariBAD, the task inference model is trained using amortized variational inference to maximize the Evidence Lower Bound (ELBO)
    \begin{equation}
        \log p(\tau) \geq \mathbb{E}_{q_\phi(z|\tau)}[\log p(\tau|z)] - D_{\mathrm{KL}}(q_\phi(z|\tau) \lVert p(z)).
    \end{equation}
    Without gradient isolation, the encoder \( q_\phi(z|\tau) \) might overfit to policy-specific features, distorting the latent belief distribution in favor of optimizing the current return rather than encoding generalizable task features. Preventing this gradient feedback ensures that the learned latent space remains well-calibrated and policy-agnostic.

    \item \textit{Reduced Instability in Policy Optimization:} reinforcement learning policy updates are known to be noisy and high-variance. If these unstable gradients influence the inference model, they can introduce non-stationarity into the belief space, making training less stable and degrading the quality of the learned task representations. Isolating the inference model from these gradients helps preserve a stable and reliable task embedding throughout learning.
\end{enumerate}

In summary, decoupling the learning dynamics between the policy and the inference model supports the formation of robust and transferable internal representations of tasks. This structural separation is critical for achieving sample-efficient adaptation in context-adaptive meta-RL, particularly in environments with high task diversity or limited supervision.

\subsection{The Action-Free Belief Model Design}\label{sec:meta-actionfree}
A central component of context-adaptive meta-reinforcement learning is the belief model, which maintains a latent representation of the current task based on interaction history. To be effective, this model must extract meaningful temporal patterns that characterize both the task dynamics and reward structures. Most frameworks incorporate some form of temporal modeling to capture these features from ongoing interactions.

For example, PEARL circumvents sequential modeling by decomposing each episode \(\tau^{\mathcal{T}_k}_{0:t} = \{(s_i, a_i, r_i, s_{i+1})\}_{i=0}^{t-1}\) into independent interaction tuples. These tuples are stored in task-specific replay buffers. During inference, \(N_c\) time-invariant context samples \(\{\mathbf{c}^{\mathcal{T}_k}_i\}_{i=1}^{N_c}\) are drawn at random from experiences with task \(\mathcal{T}_k\) and used to infer a posterior over the latent task variable. While this formulation is effective in off-policy training, it imposes two limitations. First, it implicitly assumes the Markov property across all tasks, which limits its applicability to environments with long-range temporal dependencies or partial observability. Second, PEARL allows policy gradients to backpropagate through the belief model. This coupling between the inference and control components may lead to overfitting, where the task representation becomes tailored to short-term policy performance rather than capturing generalizable features of the environment.

Alternatively, BAMDP approaches such as VariBAD treat the task context as entire sequences of interaction—either single episodes \(\tau_{0:H}\) or multi-episode sequences \(\tau_{0:H^+}\). In BAMDPs, the posterior belief over the unknown transition and reward functions is typically formulated as \(b_t = p(\mathcal{P}_S, \mathcal{P}_R \mid \tau_{0:t})\). This formalism explicitly encodes uncertainty over environment dynamics and rewards and supports exploration strategies based on belief updating.

To further decouple task inference from decision-making, and following the rationale in \Cref{sec:meta_encdec}, the posterior belief can be conditioned only on the observed states and rewards
\begin{equation}\label{eq:belief_actionfree}
    b_t^{\text{action-free}, \mathcal{T}_k} = p(\mathcal{P}_R, \mathcal{P}_S \mid \tau^{\text{action-free}, \mathcal{T}_k}_{0:t}),
\end{equation}
where the context \(\tau^{\text{action-free}, \mathcal{T}_k}_{0:t}\) omits actions entirely and may consist of multiple episodes drawn from the same task \(\mathcal{T}_k\). This context is defined as
\begin{align}\label{eq:traj_actionfree}
    \tau^{\text{action-free}, \mathcal{T}_k}_{0:t} &= \{s^1_0, r^1_1, s^1_1, \dots, r^1_H, s^1_H, \nonumber \\
     & \qquad s^2_0, r^2_1, s^2_1, \dots, r^2_H, s^2_H, \nonumber \\
     & \qquad \dots, \nonumber \\
     & \qquad s^{n_\tau}_0, r^{n_\tau}_1, s^{n_\tau}_1, \dots, r^{n_\tau}_t, s^{n_\tau}_t\},
\end{align}
where $H$ is the maximum episode length, \(n_\tau \leq n_H\) denotes the number of episodes in the context, and \(H^+ = n_H \times H\) is the maximum context horizon.

In this formulation, the belief-augmented transition and reward functions presented in \Cref{eq:bamdp_transition} and \Cref{eq:bamdp_reward} for the BAMDP are updated as
\begin{align}
    & \mathcal{P}_S^+(s_{t+1}^+ \mid s_t^+, a_{t+1}, r_{t+1}) \nonumber \\
    &= \mathcal{P}_S^+(s_{t+1}, b_{t+1} \mid s_t, a_{t+1}, r_{t+1}, b_t) \nonumber \\
    &= \mathcal{P}_S^+(s_{t+1} \mid s_t, a_{t+1}, b_t) \cdot p(b_{t+1} \mid s_{t+1}, b_t) \nonumber \\
    &= \mathbb{E}_{b_t} \left[ \mathcal{P}_S(s_{t+1} \mid s_t, a_{t+1}) \right] \cdot \delta\left(b_{t+1} = p(\mathcal{P}_R, \mathcal{P}_S \mid \tau^{\text{action-free}}_{0:t+1})\right),
    \label{eq:bamdp_transition_actionfree}
\end{align}
and
\begin{align}
    & \mathcal{P}_R^+(r_{t+1} \mid s_t^+, a_t, s_{t+1}^+) \nonumber \\
    &= \mathcal{P}_R^+(r_{t+1} \mid s_t, b_t, a_t, s_{t+1}, b_{t+1}) \nonumber \\
    &= \mathbb{E}_{b_{t+1}} \left[ \mathcal{P}_R(r_{t+1} \mid s_t, s_{t+1}) \right].
    \label{eq:bamdp_reward_actionfree}
\end{align}

This formulation naturally supports learning from demonstrations or agent-agnostic data sources, as belief updates no longer require knowledge of the agent's actions. In many robotic tasks, especially goal-oriented environments using high-level controllers, reward functions primarily depend on outcome success, rather than the precise action taken. This is particularly true in meta-RL benchmarks like Metaworld~\cite{yu2020meta}, where high-level Cartesian actions are mapped to joint motions by a robust controller. Since the influence of the action is implicitly reflected in the state transitions, this design enables the inference model to operate effectively even when actions are absent from the input sequence. Therefore, it is safe to assume that \Cref{assume:actionfree} holds for the environments in the field of interest of this paper.

\begin{assumption}\label{assume:actionfree}
Consider episodic stochastic MDP: $\mathcal{M} = (\mathcal{S}, \mathcal{A}, \mathcal{P}_R, \mathcal{P}_S, \mathcal{P}_{S_0}, \gamma, H)$ with state space $\mathcal{S}$, action space $\mathcal{A}$, transition function $\mathcal{P}_S(s'|s, a)$, reward function $\mathcal{P}_R(r|s, a, s')$, discount factor $\gamma$, and maximum trajectory length of $H$.

A Bayes-adaptive belief maintained on the dynamic functions, conditioned on a full trajectory $\tau_{0:t} = \{ s_0, a_1, r_1, \dots, s_{t-1}, a_t, r_t, s_t\}$, is defined as
\begin{equation*}
    b_t(\mathcal{P}_R, \mathcal{P}_S) = p(\mathcal{P}_R, \mathcal{P}_S \mid \tau_{0:t}).
\end{equation*}
This belief can be approximated by
\begin{equation*}
    b_t^{\text{action-free}} = p(\mathcal{P}_R, \mathcal{P}_S \mid \tau^{\text{action-free}}_{0:t}) \approx p(\mathcal{P}_R, \mathcal{P}_S \mid \tau_{0:t})
\end{equation*}
assuming a sufficiently long trajectory ($t \geq t_{\min}$), and if
\begin{equation*}
    \mathcal{P}_R(r|s,a,s') \approx \tilde{\mathcal{P}}_R(r|s_0,\dots,s_{-1},s,s'),
\end{equation*}
and,
\begin{equation*}
    \mathcal{P}_S(s'|s,a) \approx \tilde{\mathcal{P}}_S(s'|s_0,\dots,s_{-1},s).
\end{equation*}
\end{assumption}

However, this action-free design introduces several new challenges that place greater demands on the capacity and stability of the sequence model:
\begin{enumerate}[label=\Roman*.]
    \item \textit{Hierarchical and Phase-Aware Inference:} The belief model must not only infer the overall task identity but also distinguish different phases or skills within a task. This is particularly important during adaptation, where the context window is updated continuously while the belief and policy models remain frozen.

    \item \textit{Long-Term Dependency Modeling:} Removing actions reduces the amount of structured supervision available to the belief model, which now must infer latent task features from more ambiguous and flexible input. In this setting, attention-based models with autoregressive and long-range dependency capabilities are more suitable than memory-based recurrent networks.

    \item \textit{Scalability with Longer Contexts:} The action-free input sequence is more compact, allowing longer trajectories across multiple episodes to be included in the context. While this can improve task inference performance, it also necessitates robust sequence models that can prevent forgetting and capture dependencies across extended temporal horizons.
\end{enumerate}

Due to these requirements, a transformer-based belief model is adopted in the subsequent design, as detailed in the following section.

\subsection{Transformer Encoder-Decoder for Task Inference}
This section introduces the integration of a transformer encoder-decoder architecture into the belief model for the Bayes-adaptive meta-RL framework.

\subsubsection{Rotary Positional Embedding}
Positional embedding is an integral part of a transformer architecture, also present in CRAFT, to make the attention mechanism aware of the temporal relations among the tokens. For instance, in self-attention, rather than multiplying transformation matrices $\mathbf{W}_{\{q, k, v\}}$ on raw token embeddings $\mathbf{E}$, the following process is adopted in the conventional Additive positional embedding \cite{vaswani2017attention},
\begin{equation}
	\begin{aligned}
		\mathbf{q}_m &=f_q(\mathbf{E}_m, m)\\
		\mathbf{k}_n &=f_k(\mathbf{E}_n, n)\\
		\mathbf{v}_n &=f_v(\mathbf{E}_n, n),\\
	\end{aligned}
	\label{qkv}
\end{equation}
with, 
\begin{equation}
	f_{\{q, k, v\}}(\mathbf{E}_i,i):=\mathbf{W}_{\{q, k, v\}}(\mathbf{E}_i+\mathbf{p}_i).
	\label{adtv-posi}
\end{equation}
Here, $\mathbf{p}$ is commonly chosen as a sinusoidal function of both $i$ and the position of each element of the token, as in,
\begin{equation}
	\begin{cases}
		\mathbf{p}_{i,2t}&=\sin(m/10000^{2t/d})\\
		\mathbf{p}_{i,2t+1}&=\cos(m/10000^{2t/d})
	\end{cases},
	\label{pos_sin}
\end{equation}
where $i$ is the token position in the context and $t \in {0,  \dots, d/2-1}$ with $d$ representing the length of the token.

In this work, the rotary positional embedding (RoPE) \cite{su2024roformer} is used for its superiority in retaining information on the relative position of the tokens present in the dot product process of the attention mechanics. In other words, the dot product of two token embeddings is preferred to be only dependent on $m-n$, the relative position or distance between tokens $\mathbf{E}_m$ and $\mathbf{E}_n$.
\begin{equation}
	\langle f_q(\mathbf{E}_m, m),f_k(\mathbf{E}_n, n)\rangle=g(\mathbf{E}_m,\mathbf{E}_n,m-n).
	\label{roformer}
\end{equation}


The following transformation in RoPE formulation satisfies this property,

\begin{equation}
	f_{\{q, k\}}(\mathbf{E}_m, m) = \mathbf{R}^d_{\Theta, m}\mathbf{W}_{\{q, k\}}\mathbf{E}_m 
	\label{fn:rope-fqk}
\end{equation}
where, 
\begin{equation}    
	\mathbf{R}^d_{\Theta,m} = 
	\begin{pmatrix}
		\mathbf{R}^d_{\theta_1,m}&0&\cdots&0\\
		0&\mathbf{R}^d_{\theta_2,m}&\cdots&0 \\
		\vdots&\vdots&\ddots&\vdots\\
		0&0&\cdots&\mathbf{R}^d_{\theta_{d/2},m}
	\end{pmatrix}
	\label{fn:rope-RMat}
\end{equation}
is a quasi-diagonal matrix with 2-dimensional rotation matrices $\mathbf{R}^d_{\theta_i,m}$ on the diagonal,
\begin{equation}    
    \mathbf{R}^d_{\theta_i,m} = 
    \begin{pmatrix}
        \cos{m\theta_i}& -\sin{m\theta_i}\\
        \sin{m\theta_i}&\cos{m\theta_i}
    \end{pmatrix}.
    \label{fn:rope-RMat2D}
\end{equation}

The rotations are derived using similar angles as in the conventional additive sinusoidal positional embedding $\Theta = \{\theta_i=10000^{-2i/d}, i \in [0, 1, ..., d/2-1]\}$.
The dot product of two token embeddings after RoPE transformation will be

\begin{align}
	\mathbf{q}_m^{\intercal}\mathbf{k}_n &= \langle f_q(\mathbf{E}_m, m),f_k(\mathbf{E}_n, n)\rangle \nonumber \\ &= (\mathbf{R}^d_{\Theta, m}\mathbf{W}_q\mathbf{E}_m)^\intercal(\mathbf{R}^d_{\Theta, n}\mathbf{W}_k\mathbf{E}_n) \nonumber \\ &= \mathbf{E}_m^\intercal \mathbf{W}_q^\intercal R^d_{\Theta, n-m} \mathbf{W}_k \mathbf{E}_n
	\label{fn:rope-qk}
\end{align}
where $\mathbf{R}^d_{\Theta, n-m}=(\mathbf{R}^{d}_{\Theta, m})^\intercal\mathbf{R}^d_{\Theta, n}$.

Consequently, the self-attention in the general form can be rewritten as
\begin{equation}
	\operatorname{Attn}(\mathbf{Q},\mathbf{K},\mathbf{V})_m=\frac{\sum_{n=1}^{N}\operatorname{sim}(\mathbf{q}_m,\mathbf{k}_n)\mathbf{v}_n}{\sum_{n=1}^{N}\operatorname{sim}(\mathbf{q}_m, \mathbf{k}_n)}.
	\label{fn:atten-full}
\end{equation}

In the case of conventional self-attention in transformers \cite{vaswani2017attention}, $\operatorname{sim}(\mathbf{q}_m,\mathbf{k}_n)=\exp(\mathbf{q}_m^{\intercal}\mathbf{k}_n/\sqrt{d})$. The same principle applies to cross-attention as well. Incorporating this type of positional embedding makes the belief model aware of not only the positions of a state or reward, but also the distance between tokens. This will result in an improved ability to identify recurrences in tasks.

\subsubsection{Implementation in CRAFT}
\begin{figure*}[!t]
\centering
    \subfloat[]{
        \includegraphics[width=0.46\textwidth]{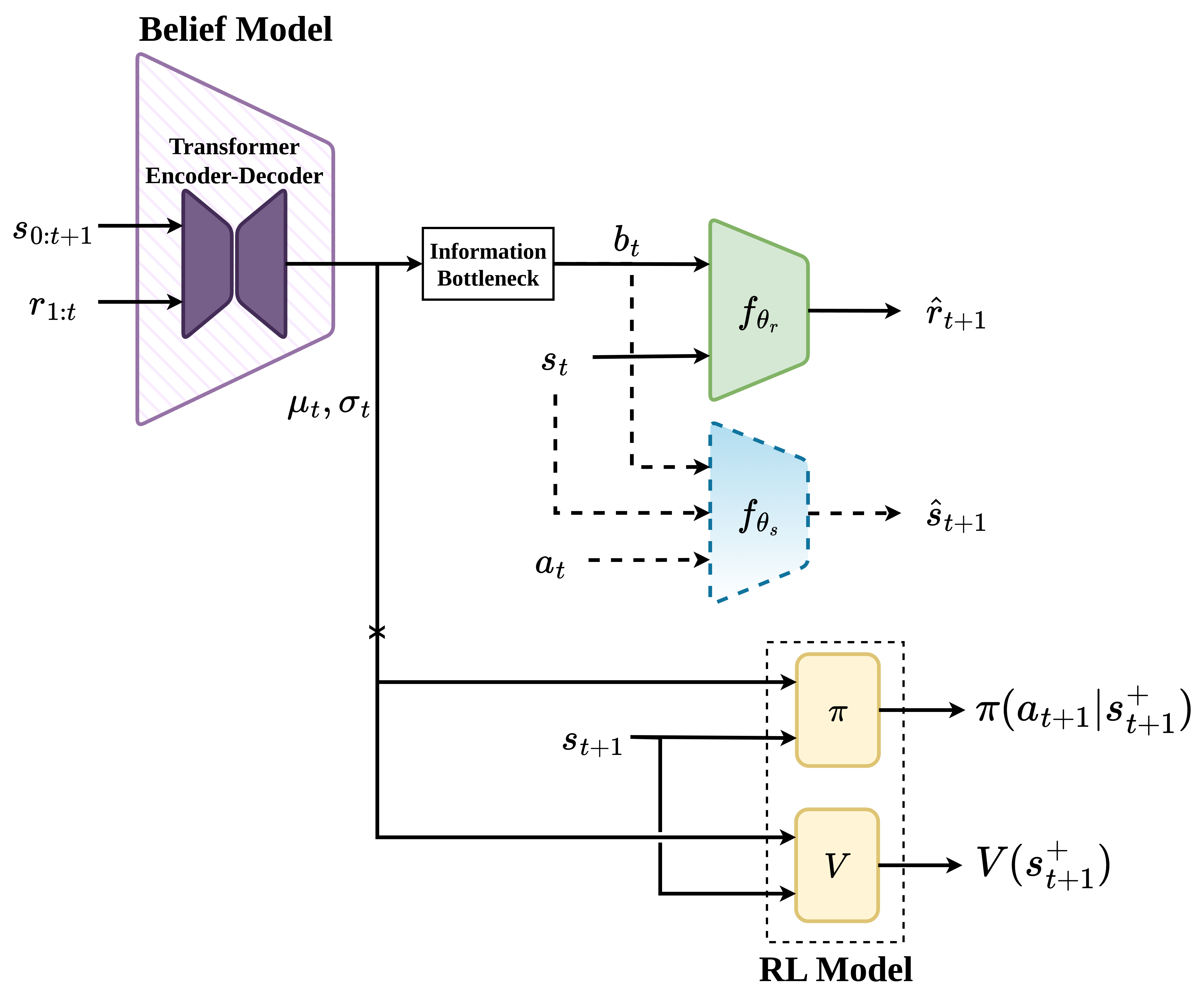}
        \label{fig:meta-overview}
    }
    \hfil
    \subfloat[]{
        \includegraphics[width=0.46\textwidth]{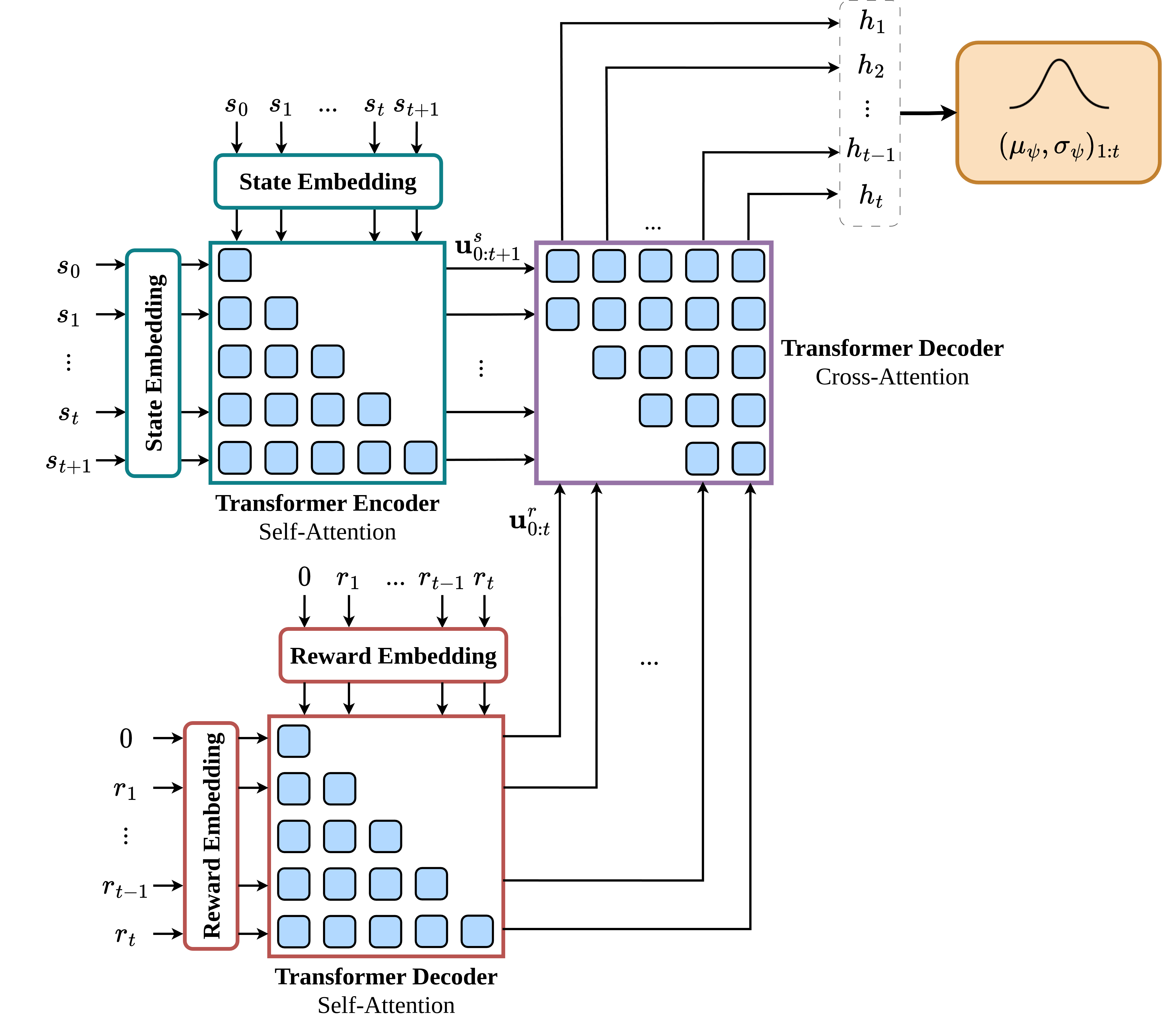}
        \label{fig:meta-tf}
    }
    \caption{\textit{Left:} General overview of the proposed action-free belief model (CRAFT), \textit{Right:} The internal structure of the belief model based on the causal transformer encoder-decoder}
\end{figure*}

The architecture shown in \Cref{fig:meta-overview} uses a transformer encoder-decoder to extract compact and informative representations from action-free trajectories, aligning with the structure of belief updates in Bayes-Adaptive Markov Decision Processes (BAMDP). Borrowing from \Cref{eq:belief_actionfree}, the posterior belief over task identity is defined as
\begin{equation*}
    b_t^{\mathcal{T}_k} = p(\mathcal{P}_R, \mathcal{P}_S | \tau^{\text{action-free}, \mathcal{T}_k}_{0:t}),
\end{equation*}
where $\tau^{\text{action-free}, \mathcal{T}_k}_{0:t}$ includes only sequences of states and rewards across one or multiple episodes. This eliminates the dependency on actions and better reflects the structure of tasks in domains like robotics, where actions often correspond to high-level commands with minimal influence on reward assignment. 


This architecture is chosen for its ability to combine two complementary inference mechanisms: the transformer encoder's self-attention extracts structural features from observed state transitions, effectively modeling the dynamic nature of the environment; meanwhile, the cross-attention in the transformer decoder captures how reward signals depend on these dynamics, defining the notion of optimality specific to the task at hand. This separation enables the belief model to represent both parametric variations (e.g., reward shaping or goal/position variations) and non-parametric variations (e.g., topology or temporal/causal structure) in a unified and flexible framework.

Excluding action tokens from the context serves two main purposes. First, it ensures that the inferred belief is disentangled from the agent's decision process, maintaining the task definition as a property of the environment rather than a byproduct of a particular policy. Second, it facilitates the use of action-free trajectories, such as demonstrations from human teleoperation or scripted controllers, for warm-starting or pre-training the inference model. This capability extends the applicability of CRAFT to scenarios where actions are unavailable or irrelevant, and opens opportunities to leverage existing datasets \cite{finn2017one, peng2018deepmimic} for task representation learning.

\paragraph{Transformer Encoder on States:}

The encoder receives a sequence \(\{s_0, s_1, \dots, s_{t+1}\}\), and embeds them into
\begin{equation*}
    \{\mathbf{E}^s_0, \mathbf{E}^s_1, \dots, \mathbf{E}^s_{t+1}\}
\end{equation*}
via a learnable transformation. The encoder then applies self-attention to capture temporal dependencies among the state tokens 
\begin{align}
    [\mathbf{u}^s_i]_{i=0:t+1} = \text{Causal SelfAttn}(&\mathbf{q}, \mathbf{k}, \mathbf{v}) \nonumber \\
    = \text{Causal SelfAttn}(&[f_q(\mathbf{E}^s_i)]_{i=0:t+1},  \nonumber \\
    & [f_k(\mathbf{E}^s_i)]_{i=0:t+1},  \nonumber \\
    & [f_v(\mathbf{E}^s_i)]_{i=0:t+1}),
\end{align}
where  $f_{\{q,k,v\}}(\mathbf{E}^s)=\mathbf{W}_{\{q,k,v\}}\mathbf{E}^s$. The inclusion of \(s_{t+1}\) ensures that each transition is fully represented up to the most recent outcome.

$\text{Causal SelfAttn}$ is the self-attention mechanism with a causal mask where each token $\mathbf{E}^s_{i+1}$, $0 \leq i \leq t$, can only attend to $\{\mathbf{E}^s_j\}_{j \leq i}$. Therefore, the undesirable attention weights will be suppressed to make the representation $\mathbf{u}^s_i$ a causal autoregressive estimation
\begin{equation}
    \mathbf{u}^s_{t+1} \approx p(s_{t+1}|s_0, \dots, s_t).
\end{equation}

The state-based encoder outputs $[\mathbf{u}^s_i]_{i=0:t+1}$ are then passed to the subsequent decoder module and will be used as queries and keys.

\paragraph{Transformer Decoder on Rewards:}

The reward sequence \(\{0, r_1, \dots, r_t\}\), after going through an embedding transformation and mapping to \(\{\mathbf{E}^r_0, \mathbf{E}^r_1, \dots, \mathbf{E}^r_t\}\), is processed with an autoregressive causal self-attention similar to the transformer encoder

\begin{align}
    [\mathbf{u}^r_i]_{i=0:t} = \text{Causal SelfAttn}(&\mathbf{q}, \mathbf{k}, \mathbf{v}) \nonumber \\
    = \text{Causal SelfAttn}(&[f_q(\mathbf{E}^r_i)]_{i=0:t},  \nonumber \\
    & [f_k(\mathbf{E}^r_i)]_{i=0:t},  \nonumber \\
    & [f_v(\mathbf{E}^r_i)]_{i=0:t}),
\end{align}
to estimate
\begin{equation}
    \mathbf{u}^r_t \approx p(r_t|0, r_1, \dots, r_{t-1}).
\end{equation}

\paragraph{Belief Formation in the Transformer Decoder:}

The decoder finally attends to both the state and the reward encodings through cross-attention, fusing dynamic information and temporal optimality structure to generate task-relevant representations 
\begin{align}
    [h_i]_{i=1:t} = \text{Causal CrossAttn}(&\mathbf{q}, \mathbf{k}, \mathbf{v}) \nonumber \\
    = \text{Causal CrossAttn}(&[f_q(\mathbf{u}^s_i)]_{i=0:t+1}, \nonumber \\
    & [f_k(\mathbf{u}^s_i)]_{i=0:t+1}, \nonumber \\
    & [f_v(\mathbf{u}^r_i)]_{i=0:t}),    
\end{align}

The causal mask here is a one-step shifted version of the ones used for the self-attention in the previous stages. This is because the cross-attention mechanism receives two sequences with different lengths, $\{\mathbf{u}^s_0, \mathbf{u}^s_1, \dots, \mathbf{u}^s_{t+1}\}$ for queries and keys, and $\{\mathbf{u}^r_0, \mathbf{u}^r_1, \dots, \mathbf{u}^r_t\}$ for values, to estimate
\begin{equation}
    h_t \approx p(r_{t+1}|s_0, s_1, \dots, s_{t+1}, 0, r_1, \dots, r_t)
\end{equation}

This design is motivated by the idea that the amortized inference of a belief relies on the approximation of the reward function, which is the most authentic definition of the tasks that share state and action spaces. It is also assumed that this can be done only with state transitions and independently of the actions. The mentioned assumption is stronger, especially in environments where outcome-based signals are dominant, such that
\begin{equation}
    b_t = p(\tilde{\mathcal{P}}_R|\tau^\text{action-free}_{0:t})
    \approx p(\tilde{\mathcal{P}}_R(r_{t+1}|s_t, s_{t+1})).
\end{equation}

These representations \( h_t \) are further compressed via a transformation parametrized by $\psi$ to a latent posterior distribution \( \mathcal{N}(\mu_\psi(h_t), \sigma_\psi(h_t)) \). All the previous stages describing the internal structure of the transformer encoder-decoder are visualized in \Cref{fig:meta-tf}.

Finally, to create an information bottleneck for regulating variational inference, the belief is sampled from this distribution
\begin{equation}
    b_t \sim \mathcal{N}(\mu_\psi(h_t), \sigma_\psi(h_t))
\end{equation}

Now, as an example, consider a trajectory with two transitions
\begin{equation}
    \tau^{\text{action-free}}_{0:2} = \{s_0, r_1, s_1, r_2, s_2\}.
\end{equation}
The encoder self-attention block receives $\{s_0, s_1, s_2\}$ and outputs $\{\mathbf{u}^s_0, \mathbf{u}^s_1, \mathbf{u}^s_2\}$, while the decoder self-attention block receives $\{0, r_1\}$ and outputs $\{\mathbf{u}^r_0, \mathbf{u}^r_1\}$. Using the latent representation $h_1$ calculated by the masked cross-attention of the transformer decoder, a decoder $f_{\theta_r}$ estimates the current reward $\hat{r}_2$ by a belief $b_1 \sim \mathcal{N}(\mu_\psi(h_1), \sigma_\psi(h_1))$ and the previous state $s_1$. Another decoder $f_{\theta_s}$ is responsible for estimating $f_{\theta_s}(s_2|s_1, a_1, b_1)$. The gradients of the loss of these estimations backpropagate through the belief model for additional guidance in the updates.


The resulting latent representation distribution, $(\mu_1, \sigma_1)$, summarizes the task-relevant latent distribution, representing belief and uncertainty, and is useful for downstream prediction and control in the RL agent.

\subsubsection{Bayesian Inference and Evidence Lower-Bound}\label{sec:meta_elbo}
The belief model outputs a posterior \( q_\phi(b_t | \tau^\text{action-free}_{0:t+1}) \) over the latent distribution \( b_t \sim p(b_t) \). The posterior $b_t$ is used to condition dynamics and reward predictors \( f_{\theta_r} \) and \( f_{\theta_s} \). The latent belief is derived from the transformer decoder’s cross-attention representations \( \{h_1, \dots, h_t\} \), transformed into Gaussian parameters:
\begin{equation}
    q_\phi(b_t | \tau_{0:t}) \sim \mathcal{N}(\mu_\psi(h_t | \tau^\text{action-free}_{0:t+1}), \sigma_\psi(h_t | \tau^\text{action-free}_{0:t+1})).
\end{equation}

The reward and transition models are conditioned on this belief and used to reconstruct the next-step reward and state, respectively. The predictive decoders are defined as:
\begin{equation}
    f_{\theta_r}(s_t, b_t) \rightarrow \hat{r}_{t+1}, \quad
    f_{\theta_s}(s_t, a_t, b_t) \rightarrow \hat{s}_{t+1}.
\end{equation}

To learn this structure, the evidence lower bound (ELBO) on the marginal trajectory likelihood is optimized:
\begin{align}
\mathcal{L}_{\text{ELBO}, t} = \mathbb{E}_{q_\phi(b_t | \tau^\text{action-free}_{0:t+1})} \left[ \log p_{\theta_r, \theta_s}(r_{t+1}, s_{t+1} | s_t, a_t, b_t) \right] & \nonumber \\
- \, \text{KL}\Bigl(q_\phi(b_t | \tau^\text{action-free}_{0:t+1}) \lVert p_{\theta_r, \theta_s}(b_t)\Bigr)&,
\end{align}
with the prior $p_{\theta_r, \theta_s}(b_t)$ set to $\mathcal{N}(0, I)$. In detail, the objective can be derived as,
\begin{align}
\nonumber \mathcal{L}_{\text{ELBO}, t} = \mathbb{E}_{q_\phi(b_t | \tau_{0:t+1})} \bigl[&
   \mathcal{L}^\text{S}_\text{Recon} + \mathcal{L}^\text{R}_\text{Recon} + \mathcal{L}_\text{Regularization}\bigr] \\ \nonumber
= \mathbb{E}_{q_\phi(b_t | \tau_{0:t+1})} \Bigl[&\log p_{\theta_r}(r_{t+1} | s_t, b_t) \\ \nonumber
&+ \log p_{\theta_s}(s_{t+1} | s_t, a_t, b_t) \\
&- \text{KL} \Bigl(q_\phi(b_t | \tau_{0:t+1}) \lVert \mathcal{N}(0,I)\Bigr)\Bigr]
\end{align}

This ensures that the latent representation \( b_t \) captures sufficient information to predict both reward and state transitions, thereby supporting fast and generalizable adaptation. By optimizing this objective, the model learns to infer latent task features from action-free contexts and propagate uncertainty into predictive modules, in alignment with BAMDP principles.

The combination of information bottleneck regularization and cross-attentive encoding supports modular design and interpretability. The separation between inference and decision-making also improves policy robustness across diverse task distributions.

\section{Related Work}\label{sec:meta_related}

\subsection{Context-Adaptive Meta-Reinforcement Learning}
Context-adaptive meta-RL enables agents to adapt to new tasks by conditioning policies on an inferred latent representation of the \textit{context}, which may include transitions or other task-related information. Rather than performing gradient-based updates at test time, these methods learn a mapping from trajectory data to a latent task embedding used for fast adaptation. PEARL \cite{rakelly2019efficient} is a representative framework that infers a variational latent for conditioning an off-policy actor–critic agent.

Among existing strategies, state augmentation, i.e., concatenating the learned representation with the observation, offers simplicity and robustness. Unlike hypernetwork-based approaches \cite{beck2023hypernetworks}, which tune policy parameters directly from the context, state augmentation preserves the Markov property and scales effectively with task complexity \cite{beck2023survey}. These context-conditioned policies perform well in continuous control and integrate smoothly with both on- and off-policy methods. Moreover, separating inference from policy optimization improves interpretability and modularity, allowing structured priors to be incorporated under uncertainty \cite{humplik2019taskinference}. Such context-adaptive designs form the foundation for more advanced Bayes-adaptive meta-RL frameworks that build on explicit task inference for generalization.

\subsection{Off-Policy Context-Adaptive Meta-RL}
Several off-policy meta-RL algorithms incorporate context embeddings.  
PEARL \cite{rakelly2019efficient} learns a latent variable $z$ to condition the policy $\pi_\theta(a|s,z)$ using an inference network $q_\phi(z|\mathbf{c})$, where $\mathbf{c}$ is a set of task-specific transitions,
\begin{equation}
\mathbf{c}^{\mathcal{T}_{\mathrm{i}}}_n={(s,a,r,s^\prime)}^{\mathcal{T}_i}_n.
\end{equation}
At inference, $q_\phi(z^\mathcal{T}|\mathbf{c}^\mathcal{T})$ updates as new transitions are observed, allowing $\pi_\theta$ to adapt rapidly. Its main drawback lies in neglecting temporal structure, limiting application to hierarchical or partially observable domains.

SiMPL \cite{nam2022skill} extends PEARL for long-horizon, sparse-reward tasks using offline data. It extracts reusable skills and a skill prior, then meta-trains a high-level policy to compose these skills into longer behaviors. The high-level policy samples latent skill vectors $z$ conditioning a low-level controller, enabling adaptation without reward or task labels. Experiments on continuous control tasks show significant gains over PEARL.

TIGR \cite{bing2023tigr} further generalizes Bayesian meta-RL to the off-policy setting using Gaussian variational autoencoders and gated recurrent units to capture multi-modal task representations. Although such PEARL-based methods handle parametric variations effectively, they struggle with non-parametric or non-stationary task distributions, relying on structured priors and fixed latent spaces that limit flexibility.

\subsection{Transformers in Meta-RL}
Meta-learning algorithms for Bayesian meta-RL include recurrent, gradient-based, and transformer-based models. Recurrent methods like RL$^2$ encode experience through hidden states but suffer from information bottlenecks over long meta-horizons $H^+$. Gradient-based learners such as MAML enable fast adaptation via a few gradient steps but are computationally heavy and often fail to generalize across diverse or hierarchical tasks.

Transformers reframe RL as sequence modeling. Decision Transformer (DT) \cite{chen2021decision} models trajectories autoregressively to output actions conditioned on desired returns, removing the need for explicit value functions. Online Decision Transformer (ODT) \cite{zheng2022online} adds online fine-tuning, and Prompt-DT \cite{xu2022prompting} incorporates short demonstration prompts as task labels for few-shot generalization. Though positioned within multi-task RL, Prompt-DT exemplifies transformer-driven generalizable policy generation.

Architectures such as TrMRL \cite{melo2022transformers} and HTrMRL \cite{shala2024hierarchical} apply self-attention for parallel sequence processing, improving efficiency and long-term memory. TrMRL introduces episodic memory reinstatement for faster adaptation, while HTrMRL captures hierarchical dependencies across and within episodes. AMAGO \cite{grigsby2023amago} combines off-policy learning with long-sequence transformers to mitigate memory and horizon limitations in sparse-reward, goal-conditioned tasks. Although effective for large-scale parallel training, AMAGO does not explicitly address task uncertainty through amortized inference, focusing instead on goal relabeling.

Overall, transformer-based meta-RL improves scalability, long-range credit assignment, and uncertainty modeling compared to recurrent and gradient-based methods. Architectures such as TrMRL, HTrMRL, and AMAGO demonstrate the growing potential of sequence models for efficient adaptation and generalization in robotic RL.

\section{Experiments}
\label{sec:meta_finding}
This section presents experimental results for the CRAFT belief inference model, built upon a transformer encoder-decoder architecture and designed for adaptive meta-RL dealing with multi-task robotic environments. The experiments evaluate the ability of the model to infer task structure and adapt policies across diverse tasks in a conventional simulation benchmark designed for such an objective. The section is organized into environment setup, implementation details, performance results, qualitative insights, and discussion.

\subsection{Environments and Training Pipeline}
\label{sec:meta_exp}

CRAFT is evaluated using the MetaWorld benchmark~\cite{yu2020meta}, a simulated open-source suite of 50 robotic manipulation tasks designed to train and evaluate meta-RL and multi-task RL algorithms. Unlike previous benchmarks that focus on narrow parametric variations, such as different positions or velocities, MetaWorld not only includes parametric variations but also offers diverse non-parametric changes across tasks. This benchmark also enables generalization to entirely new held-out tasks. This feature makes it a suitable platform to evaluate general-purpose adaptation, particularly before deployment in physical systems.


From this suite, the ML-10 scenario is adopted for the following implementation. ML-10 provides $n_\text{train}=10$ training tasks and $n_\text{test}=5$ disjoint test tasks, and each task comes with $n_\text{variations}=50$ different parametric variants. All these fifteen tasks differ entirely in terms of entities, the corresponding dynamics, and objectives. In each epoch of meta-training, a subset of training environments is sampled $\mathcal{T}^k \sim \mathcal{T}^\text{train}$, regardless of their identity or parametric configuration. Then, parallelized across the tasks in $\mathcal{T}^k$, a series of consecutive $n_H$ episodes of interaction is executed synchronously in each environment. The goal here is to accumulate contextual raw information in each meta-episode and update the belief online, following the Bayes-adaptive exploration approach in \Cref{eq:belief_actionfree,eq:traj_actionfree}.

\begin{figure}[ht!]
    \centering
    \includegraphics[width=\columnwidth]{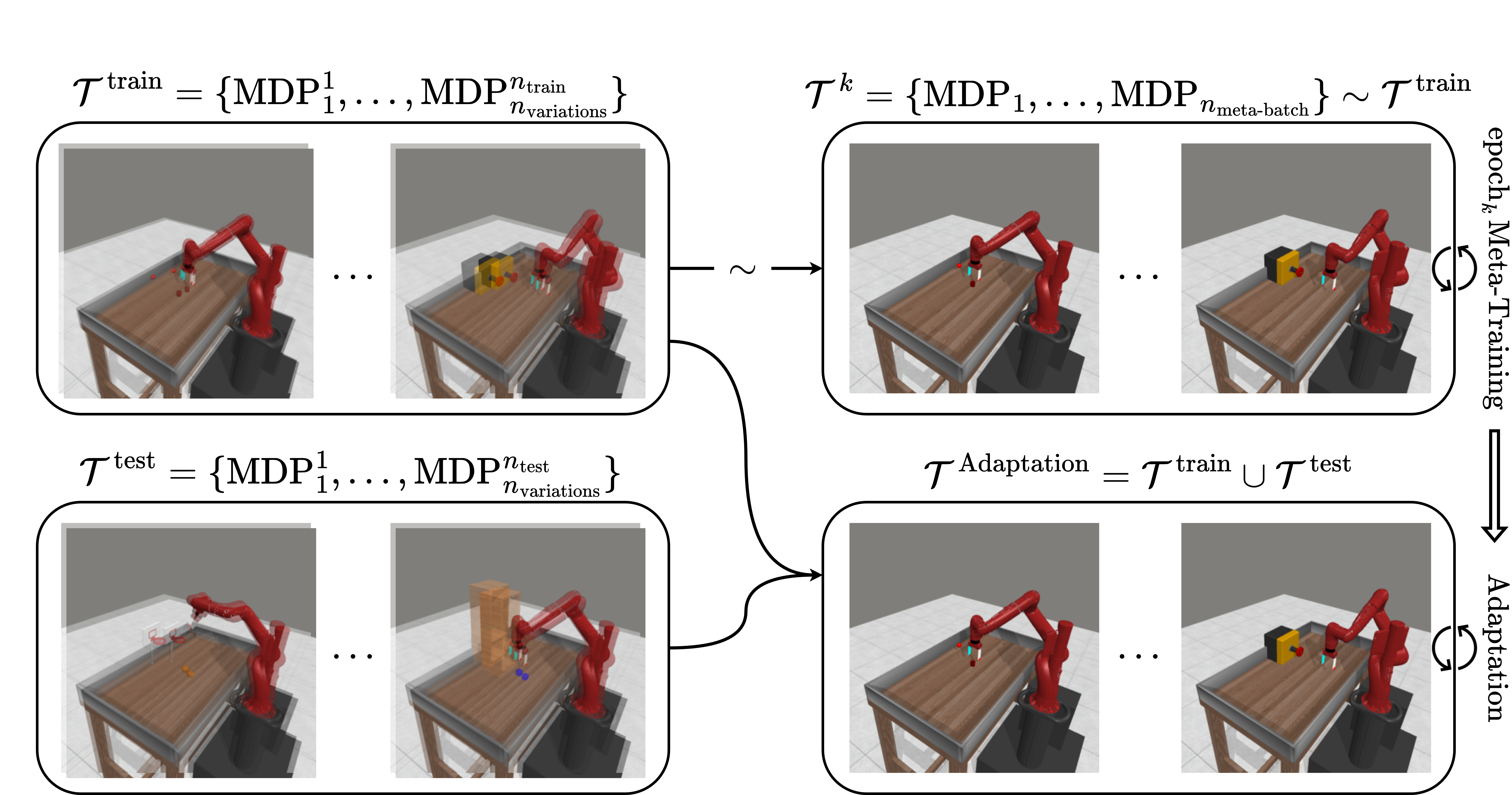}
    \caption{Distribution of task variations in MetaWorld ML-10}
    \label{fig:meta-sets}
\end{figure}

Following some isolated update iterations of the RL agent, i.e., actor and critic networks, evaluation is then performed on all variants of both the training and test sets $\mathcal{T}^\text{train} \cup \mathcal{T}^\text{test}$. \Cref{fig:meta-sets} illustrates how the environments are sampled in meta-training and adaptation phases. It should be noted that the RL updates need a separate buffer for keeping all the action and belief information in the interactions. To quantify adaptation to each task, the average return and success rate of all its variations are computed. This separation enables detailed analysis of generalization to familiar and unfamiliar environments under both identity and parametric diversity.

In addition to computing the performance metrics over each $n_H$-episode epoch, the success rate and return of the final episode in each sequence are also logged. This provides a direct measure of how accumulated context over $n_H$ episodes contributes to performance. By comparing the statistics from the average episodes and final episodes within a block, the effect of belief update on exploration can be quantified.

The implementation in this experiment uses an on-policy PPO agent \cite{schulman2017proximal}. As pointed out previously, the policy uses an online buffer that stores the most recent $n_H$ episodes. After each meta-episode, the policy is updated for $n^\pi_{\text{steps}}$ iterations. In parallel, an offline buffer keeps contextual trajectories of interactions with the training set $\tau^{\mathcal{T}_k}_{0:H^+}$ from all epochs. The belief model is updated after every $n_H$ episodes using this stored data. Again, it is worth mentioning that the belief model is completely isolated from the PPO policy, i.e., there is no shared gradient flow. This design choice avoids instability from entangled updates and ensures modularity.

CRAFT is compared against several on-policy meta-RL baselines. These include MAML and RL$^2$, widely known meta-RL baselines, as well as VariBAD, the Bayes-adaptive meta-RL method with memory-based belief model, and SDVT \cite{lee2023sdvt}. SDVT aimed to extend VariBAD by improving generalization to non-parametric task variations. To do so, SDVT proposed a more rigorous task decomposition into subtasks by learning a Gaussian mixture VAE updated by minimizing categorical and occupancy losses, and a virtual training strategy. Hypothetical latent representations, i.e., subtask composition in their work, are introduced to the RL agent in virtual training.

\subsection{Training Implementation Details}
\label{sec:meta_train}
To support reproducibility and clarity, the full training framework is outlined in Algorithm \ref{algo:meta}. It summarizes the rollout process and shows how the belief model and PPO policy are trained concurrently but independently, using separate buffers populated during meta-rollouts. In regard to baselines, for MAML and RL$^2$, the implementations from \citep{mclean2025meta} are used\footnote{\url{https://github.com/rainx0r/metaworld-algorithms}}. The implementations from the SDVT paper are also adopted for their algorithm and VariBAD\footnote{\url{https://github.com/suyoung-lee/SDVT}}. Finally, the ML-10 benchmark is borrowed from the well-maintained Metaworld simulation repository\footnote{\url{https://github.com/Farama-Foundation/Metaworld}}. 

\begin{algorithm*}[ht!]
\caption{Belief Inference from Action-Free Transformer Model}
\label{algo:meta}
\KwIn{Training and test tasks $\mathcal{T}^\text{train}, \mathcal{T}^\text{test}$\; Transformer: encoder $f^s$, decoder $f^r$, $n_\text{head}$ attention heads, $n_\text{block}$ transformer blocks, $n_\text{FFN}$ units in feedforward network, inference head $f^\psi$, reward and state predictors $f_{\theta_r}, f_{\theta_s}$\; Rollout Horizon $H$, $n_H$ rollouts in meta-episode, policy $\pi_\zeta$, value function $V_\nu$\; Belief buffer $\mathcal{B}_\text{belief}$, rollout buffer $\mathcal{B}_\text{RL}$, ELBO batch size $n_\text{ELBO}$}

\For{\text{each meta-episode} $k = 1$ to $n_\text{meta}$}{
    Sample tasks from the training tasks $\mathcal{T}^\text{meta-batch} \sim \mathcal{T}^\text{train}$ \;
    Reset hidden state $h_0$, $z_0=(\mu_\psi(h_0), \sigma_\psi(h_0))$, $\mathcal{B}_{\text{RL}}$, and $\tau^{\text{action-free}}$ \;

    \For{$t = 0$ to $n_H \times H - 1$}{
        \If{$t \mod H = 0$}{
            Reset tasks and sample $s_0 \sim \mathcal{P}_{S_0}^\text{meta-batch}(.)$\;
            Append $(\varnothing, s_0)$ to $\tau^{\text{action-free}}$ \;
            $s_t \leftarrow s_0$
        }
        
        Sample actions $a_t \sim \pi_\zeta(.|s_t, z_t)$
        Execute the actions, observe next states $s_{t+1}$, and receive rewards $r_{t+1}$\;

        Store $(s_t, z_t, r_{t+1}, a_{t+1})$ in $\mathcal{B}_\text{RL}$ \;
        
        Append $(r_{t+1}, s_{t+1})$ to $\tau^{\text{action-free}}$ \;
        
        Embed $[s_0, \dots, s_{t+1}] \subset \tau^\text{action-free}$ and apply Rotary Positional Encoding $\rightarrow [\mathbf{E'}_0^s, \dots, \mathbf{E'}_{t+1}^s]$ \;
        Apply causal self-attention: \\$\mathbf{u}_t^s = \text{FFN(Multi-Head(CausalSelfAttn}(f^s_q(\mathbf{E'}^s), f^s_k(\mathbf{E'}^s), f^s_v(\mathbf{E'}^s))))$ \;

        Embed $[0, r_1, \dots, r_t] \subset \tau^\text{action-free}$ and apply Rotary Positional Encoding $\rightarrow [\mathbf{E}_0^r, \dots, \mathbf{E}_t^r]$ \;
        Apply causal self-attention: \\$\mathbf{u}_t^r = \text{FFN(Multi-Head(CausalSelfAttn}(f^r_q(\mathbf{E'}^r), f^r_k(\mathbf{E'}^r), f^r_v(\mathbf{E'}^r)))))$ \;

        Fuse via causal cross-attention: \\$h_t = \text{FFN(Multi-Head(CausalCrossAttn}(f^r_q(\mathbf{u}^s), f^r_k(\mathbf{u}^s), f^r_v(\mathbf{u}^r)))))$ \;
        Update $z_t=(\mu_\psi(h_t), \sigma_\psi(h_t))$ \;
        Sample $b_t \sim \mathcal{N}(\mu_\psi(h_t), \sigma_\psi(h_t))$ \;

    }
    Store $\tau^\text{action-free}_{0:t+1}$ in $\mathcal{B}_\text{belief}$\;
    
    \tcc{Update}
    Sample trajectories $\tau^{1:n_\text{ELBO}} \sim \mathcal{B}_\text{belief}$ and infer belief $\{\{b_t(\tau^i_{0:t+1})\}_{t=0:H^+-1}\}^{i=1:n_\text{ELBO}}$\;
    Optimize $\mathcal{L}_\text{ELBO}=\sum_{\tau}\sum_{t=0}^{H^+-1} \mathcal{L}_{\text{ELBO}, t}$ with:\\
    $\mathcal{L}_{\text{ELBO}, t} = 
    \mathbb{E}_{q_\phi(b_t | \tau_{0:t+1})} \left[
    \beta^\text{S}\mathcal{L}^\text{S}_\text{Recon} + \beta^\text{R}\mathcal{L}^\text{R}_\text{Recon} + \beta^\text{KL}\mathcal{L}_\text{Regularization}\right]$\;

    Optimize PPO $\mathcal{L}_\pi$, $\mathcal{L}_V$ to update $\pi_\zeta, V_\nu$\;
}
\end{algorithm*}

The detailed hyperparameter configuration used throughout the experiments is provided in \Cref{tab:meta_hyperparams}, reporting relevant configurations for the belief inference module, the transformer architecture, and the reinforcement learning agent.

\begin{table}[hb!]
\centering
\caption{Action-Free Transformer Belief Model and RL Hyperparameters}
\label{tab:meta_hyperparams}
\begin{adjustbox}{width=0.9\columnwidth}
\begin{tabular}{@{}llcl@{}}
\toprule
\textbf{Category} & \textbf{Description} & \textbf{Value} \\

\midrule
\multirow{10}{*}{\textbf{Belief Model}} 
& Optimizer & Adam \\
& Learning rate & \(1\times 10^{-3}\) \\
& Buffer size (meta-episodes) & 1000 \\
& $n_\text{ELBO}$ & 10\\
& ELBO KL coefficient ($\beta^\text{KL}$) & 0.1 \\
& ELBO reward coefficient ($\beta^\text{R}$) & 10 \\
& ELBO transition coefficient ($\beta^\text{S}$) & 200 \\
& Number of updates per epoch & 10 \\
& Reward reconstruction objective & MSE \\
& State reconstruction objective & MSE \\

\midrule
\multirow{9}{*}{\textbf{Transformer}} 
& $d_q, d_k, d_v$ & 256 \\
& $n_\text{block}$ & 1 \\
& $n_\text{head}$ & 4 \\
& $n_\text{FFN}$ & 512 \\
& State embedding size & $32$ \\
& Reward embedding size & $16$ \\
& Latent dimension ($d_h$) & 5 \\
& Reward decoder network & [64, 64, 32] \\
& State decoder network & [64, 64, 32] \\

\midrule
\multirow{10}{*}{\textbf{PPO}} 
& Optimizer & Adam \\
& Learning rate & \(7\times 10^{-4}\) \\
& $\gamma$ & 0.99 \\
& Clip factor & 0.1 \\
& $\tau$ & 0.9 \\
& Entropy Coefficient & \(1\mathrm{e}{-3}\) \\
& Mini-batch size & 10 \\
& Architecture & $[256,256]$ \\
& Policy activation & $\tanh$ \\
& Policy state embedding size & $64$ \\
& Policy belief embedding size & $64$ \\

\bottomrule
\end{tabular}
\end{adjustbox}
\end{table}

\subsection{Results and Analysis}
\label{sec:meta_results}
Here, the findings of the aforementioned experiments are explained in two categories of quantitative and qualitative comparisons.

\subsubsection{Quantitative Findings}
\begin{figure}[ht!]
    \centering
    \includegraphics[width=\columnwidth]{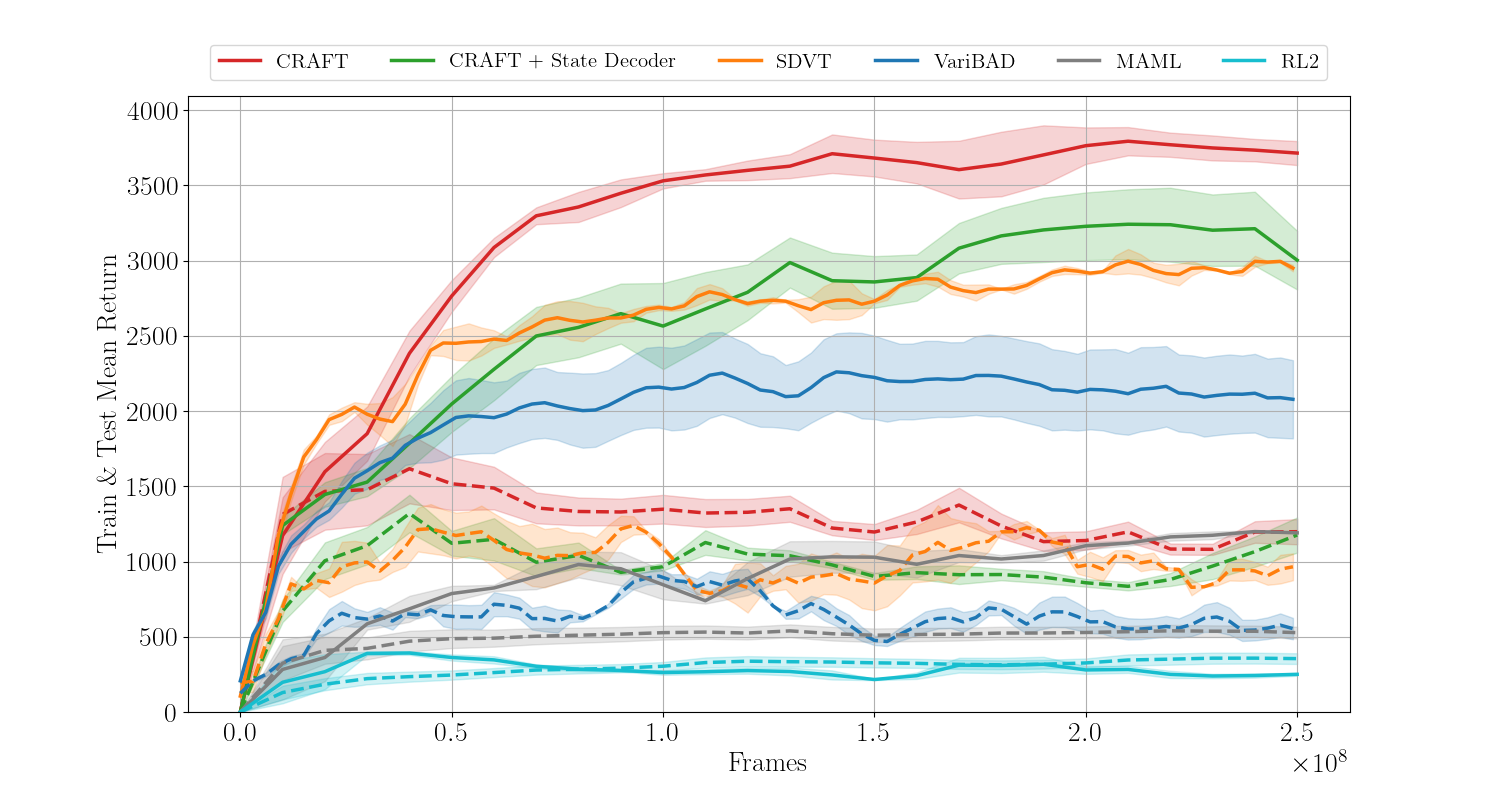}
    \caption{Average return of different methods in evaluation on training and test environments of the MetaWorld ML-10}
    \label{fig:meta-rplot}
\end{figure}

\begin{figure}[ht!]
    \centering
    \includegraphics[width=\columnwidth]{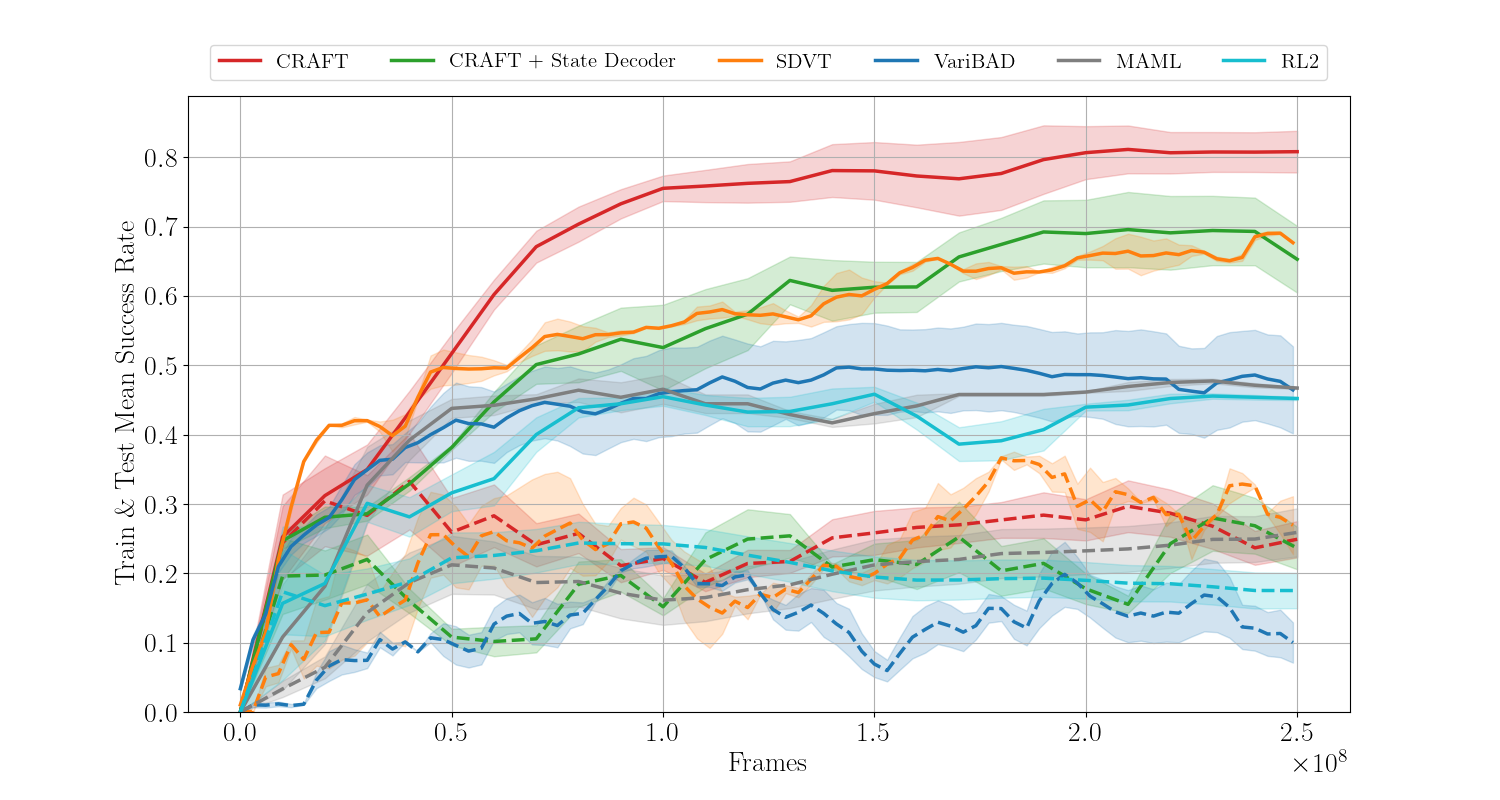}
    \caption{Average success rate of different methods in evaluation on training and test environments of the MetaWorld ML-10}
    \label{fig:meta-splot}
\end{figure}

The quantitative evaluation focuses on two performance metrics, return and success rate, across all task variations in the MetaWorld ML-10 benchmark. The average values of these metrics are plotted against the number of training interaction frames in \Cref{fig:meta-rplot} and \Cref{fig:meta-splot}. Results are reported separately for training and test environments, ensuring that generalization can be meaningfully assessed. \Cref{tab:meta-return} and \Cref{tab:meta-success} present more detailed numerical findings, showing the post-training generalization performance in each environment for CRAFT and other Bayes-adaptive baselines.

The transformer-based belief model consistently outperforms the baseline algorithms across many environments. The performance advantage is most pronounced in training environments. Although CRAFT uses less information and a less constrained model, it achieves comparable, and in some cases even better, results on the test environments that involve unfamiliar dynamics and temporal dependencies.

Another finding is that in the early stages of training, when the belief model is immature, the performance of memory-based methods is better compared to CRAFT. However, after the first few evaluations, the action-free transformer model begins to produce meaningful guidance for the context-adaptive PPO more effectively than the baselines. This leads to higher success rates and returns in later stages, where it seems the action-free transformer models have not yet even reached their full potential. This can be attributed to the complexity of training transformer models; they are slightly slower but eventually effective, even when consisting of only a small number of blocks.

\begin{table*}[ht]
\centering
\caption{Return comparison among methods across MetaWorld ML-10 tasks}
\label{tab:meta-return}
\begin{adjustbox}{width=0.8\textwidth}
\begin{tabular}{lcccc}
\toprule
\textbf{Task} & \textbf{CRAFT} & \textbf{CRAFT + State Decoder} & \textbf{SDVT} & \textbf{VariBAD} \\
\midrule
Reach & $\mathbf{3696.9\pm136.0}$ & $3641.1\pm162.0$    & $3016.8\pm216.5$ & $3588.7\pm220.5$   \\
Push & $\mathbf{3000.2\pm833.0}$ & $2360.2\pm783.5$    & $2445.6\pm250.2$ & $889.7\pm614.1$    \\
Pick-Place & $\mathbf{1528.7\pm434.0}$ & $1029.1\pm370.9$    & $1180.0\pm4.6$   & $382.7\pm266.4$    \\
Door Open & $3705.0\pm770.4$ & $3625.8\pm477.7$    & $\mathbf{4339.0\pm96.6}$  & $2797.8\pm1082.3$  \\
Drawer Close & $4734.1\pm97.0$  & $4749.0\pm64.7$     & $\mathbf{4777.9\pm9.4}$   & $4694.9\pm3.1$     \\
Button Press Topdown & $2977.0\pm620.2$ & $2665.0\pm244.0$    & $\mathbf{3134.4\pm270.1}$ & $1993.9\pm613.4$   \\
Peg Insert Side & $\mathbf{2455.5\pm722.4}$ & $1381.9\pm467.0$    & $1451.6\pm172.0$ & $774.6\pm540.7$    \\
Window Open & $3692.2\pm833.5$ & $\mathbf{4289.6\pm43.4}$     & $4219.9\pm101.7$ & $3081.2\pm780.5$   \\
Sweep & $\mathbf{3332.7\pm921.4}$ & $2293.3\pm735.3$    & $2820.2\pm227.0$ & $1527.8\pm1050.8$  \\
Basketball & $\mathbf{2545.5\pm781.8}$ & $1346.4\pm502.7$    & $1722.6\pm110.9$ & $868.9\pm609.2$    \\
\midrule
Drawer Open & $2143.4\pm313.3$ & $2043.6\pm40.3$     & $\mathbf{2144.3\pm84.0}$   & $1589.5\pm352.7$  \\
Door Close & $791.0\pm359.2$  & $318.3\pm72.6$      & $\mathbf{1185.7\pm53.0}$   & $261.8\pm4.8$     \\
Shelf Place & $\mathbf{409.9\pm123.3}$  & $232.0\pm77.4$      & $354.3\pm50.3$    & $61.1\pm43.2$     \\
Sweep Into & $\mathbf{893.2\pm324.3}$  & $504.5\pm141.1$     & $754.8\pm145.7$   & $359.0\pm223.2$   \\
Lever Pull & $\mathbf{346.7\pm39.0}$   & $324.6\pm18.5$      & $292.6\pm26.9$    & $337.5\pm2.9$     \\
\midrule
Train Avg. & $\mathbf{3166.8}$       &       $2738.1$    &     $2910.8$        &     $2060.03$ \\
Test Avg. & $916.82$       &       $684.6$     &     $\mathbf{946.4}$         &     $521.8$   \\
Total Avg. & $\mathbf{2416.8}$       &       $2053.6$    &     $2256.0$        &     $1547.3$  \\
\bottomrule
\end{tabular}
\end{adjustbox}
\end{table*}

\begin{table*}[ht]
\centering
\caption{Success rate comparison among methods across MetaWorld ML-10 tasks}
\label{tab:meta-success}
\begin{adjustbox}{width=0.8\textwidth}
\begin{tabular}{lcccc}
\toprule
\textbf{Task} & \textbf{CRAFT} & \textbf{CRAFT + State Decoder} & \textbf{SDVT} & \textbf{VariBAD} \\
\midrule
Reach & $0.401 \pm 0.073$ & $0.413 \pm 0.061$ & $0.363 \pm 0.002$ & $\mathbf{0.487 \pm 0.015}$ \\
Push & $\mathbf{0.555 \pm 0.158}$ & $0.491 \pm 0.165$ & $0.497 \pm 0.078$ & $0.150 \pm 0.106$ \\
Pick-Place & $0.394 \pm 0.116$ & $0.227 \pm 0.101$ & $\mathbf{0.410 \pm 0.064}$ & $0.078 \pm 0.054$ \\
Door Open & $0.798 \pm 0.223$ & $0.812 \pm 0.148$ & $\mathbf{0.997 \pm 0.002}$ & $0.501 \pm 0.353$ \\
Drawer Close & $\mathbf{1.000 \pm 0.000}$ & $0.995 \pm 0.004$ & $\mathbf{1.000 \pm 0.000}$ & $\mathbf{1.000 \pm 0.000}$ \\
Button Press Topdown  & $0.918 \pm 0.089$ & $\mathbf{0.946 \pm 0.033}$ & $0.943 \pm 0.035$ & $0.795 \pm 0.145$ \\
Peg Insert Side & $\mathbf{0.530 \pm 0.171}$ & $0.175 \pm 0.061$ & $0.210 \pm 0.021$ & $0.103 \pm 0.073$ \\
Window Open & $0.821 \pm 0.194$ & $0.990 \pm 0.011$ & $\mathbf{1.000 \pm 0.000}$ & $0.797 \pm 0.134$ \\
Sweep & $\mathbf{0.774 \pm 0.217}$ & $0.561 \pm 0.187$ & $0.713 \pm 0.071$ & $0.377 \pm 0.266$ \\
Basketball & $\mathbf{0.664 \pm 0.196}$ & $0.238 \pm 0.163$ & $0.503 \pm 0.016$ & $0.170 \pm 0.120$ \\
\midrule
Drawer Open & $0.428 \pm 0.203$ & $\mathbf{0.529 \pm 0.005}$ & $0.427 \pm 0.071$ & $0.261 \pm 0.122$ \\
Door Close & $0.200 \pm 0.133$ & $0.014 \pm 0.014$ & $\mathbf{0.223 \pm 0.071}$ & $0.020 \pm 0.004$ \\
Shelf Place & $0.007 \pm 0.007$ & $0.000 \pm 0.000$ & $\mathbf{0.013 \pm 0.009}$ & $0.000 \pm 0.000$ \\
Sweep Into & $0.303 \pm 0.134$ & $0.364 \pm 0.121$ & $\mathbf{0.523 \pm 0.059}$ & $0.213 \pm 0.151$ \\
Lever Pull & $\mathbf{0.007 \pm 0.007}$ & $0.004 \pm 0.004$ & $0.000 \pm 0.000$ & $0.000 \pm 0.000$ \\
\midrule
Train Avg. & $\mathbf{0.690}$ & $0.580$ & $0.660$ & $0.450$ \\
Test Avg. & $0.190$ & $0.180$ & $\mathbf{0.237}$ & $0.100$ \\
Total Avg. & $0.520$ & $0.450$ & $\mathbf{0.530}$ & $0.330$ \\
\bottomrule
\end{tabular}
\end{adjustbox}
\end{table*}

\begin{figure}[ht!]
    \centering
    \includegraphics[width=\columnwidth]{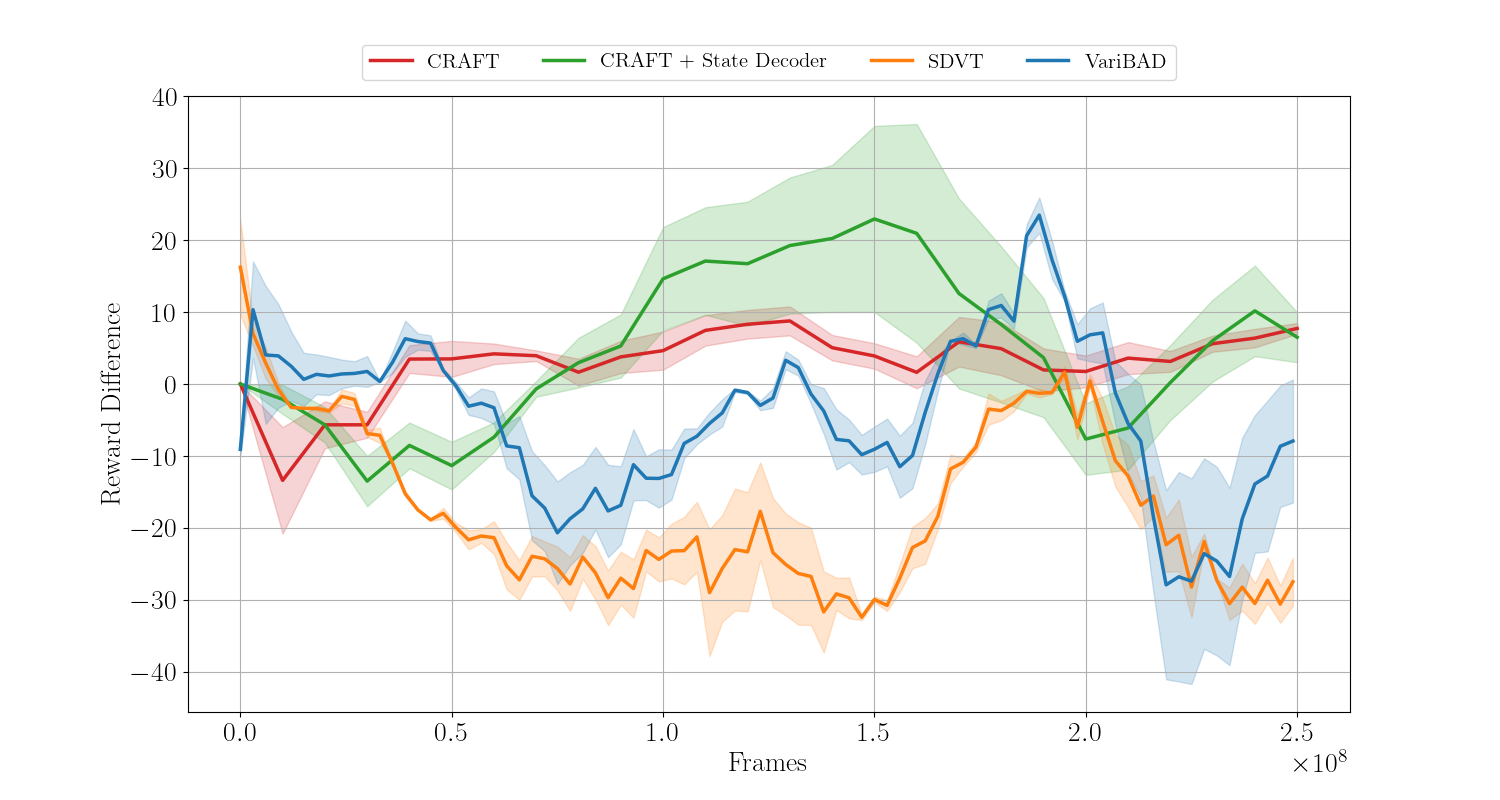}
    \caption{The difference between the return in the last episode of an adaptation meta-episode and the average ($R_f - \overline{R}_{H^+}$)}
    \label{fig:meta-rfvr}
\end{figure}

\begin{figure}[ht!]
    \centering
    \includegraphics[width=\columnwidth]{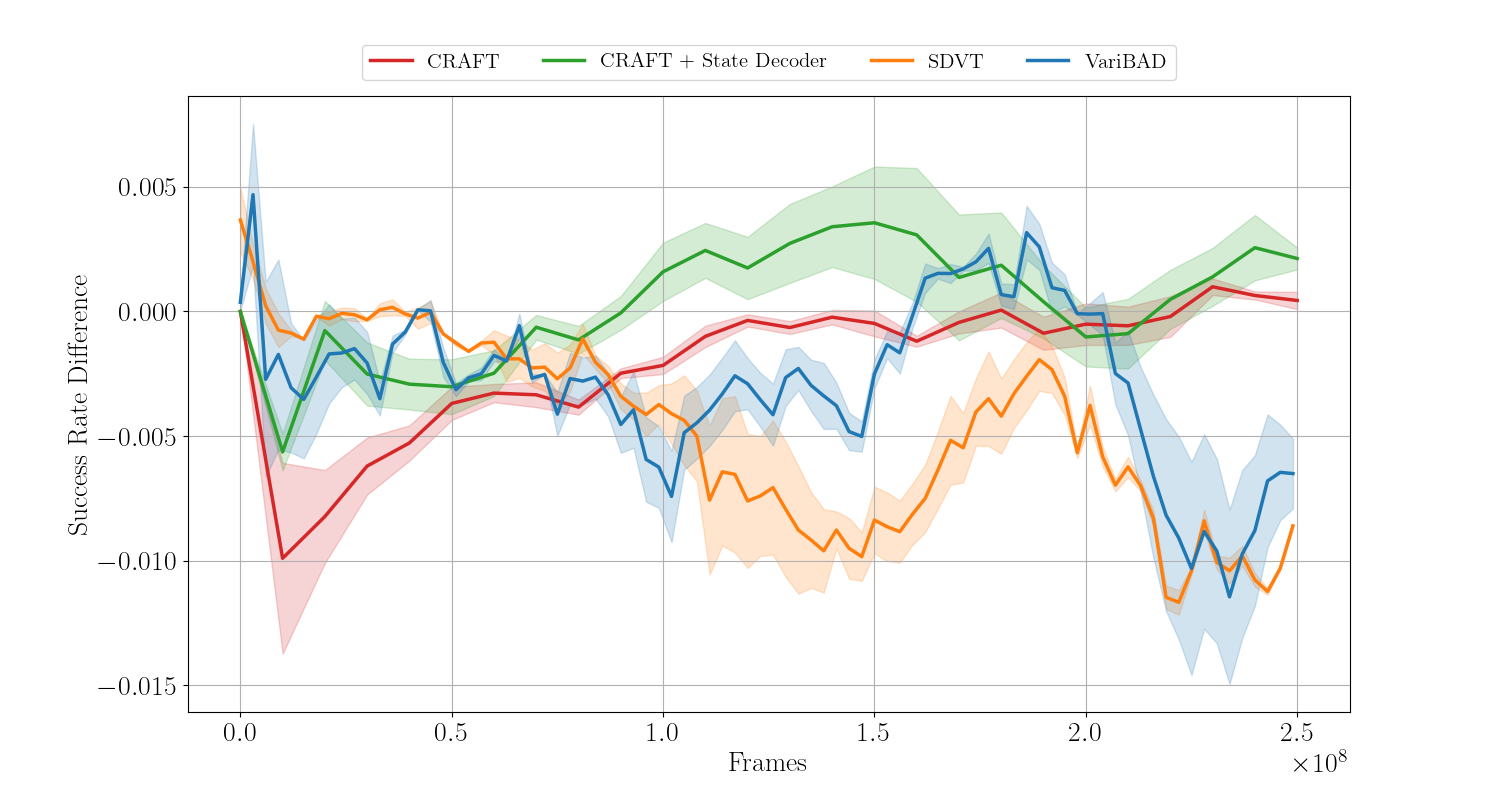}
    \caption{The difference between the success rate in the last episode of an adaptation meta-episode and the average ($S_f-\overline{S}_{H^+}$)}
    \label{fig:meta-sfvs}
\end{figure}

During many stages of training, CRAFT achieves higher final-episode return and success compared to the mean over the $n_H$ episodes in a meta-episode, denoted as $R_f - \overline{R}_{H^+}$ and $S_f-\overline{S}_{H^+}$, respectively. The corresponding values are plotted against the number of meta-training interaction frames in \Cref{fig:meta-rfvr,fig:meta-sfvs}. As can be seen, this relative advantage remains in the later stages of training for the action-free methods, suggesting that the policy can still benefit from continued exploration and training. This indicates that accumulated experience within the meta-episode contributes meaningfully to improved performance, reflecting effective belief updates in the BAMDP framework. In contrast, other baseline methods show weaker improvement—or even degradation in some cases—between early and final episodes. The ability of transformer-based models to better extract information from long contexts compared to memory-based sequence models contributes to this result. It should also be noted that CRAFT, particularly when a state decoder is included during training, shows the most noticeable benefit during intermediate training stages, when the belief model is still developing and the policy is actively exploring. This further supports the value of gradient isolation in transformer-based context encoder training, which enables informative task inference independently and provides a strong prior for the adaptation process.

\subsubsection{Qualitative Findings}
\begin{figure}[ht!]
    \centering
    \includegraphics[width=\columnwidth]{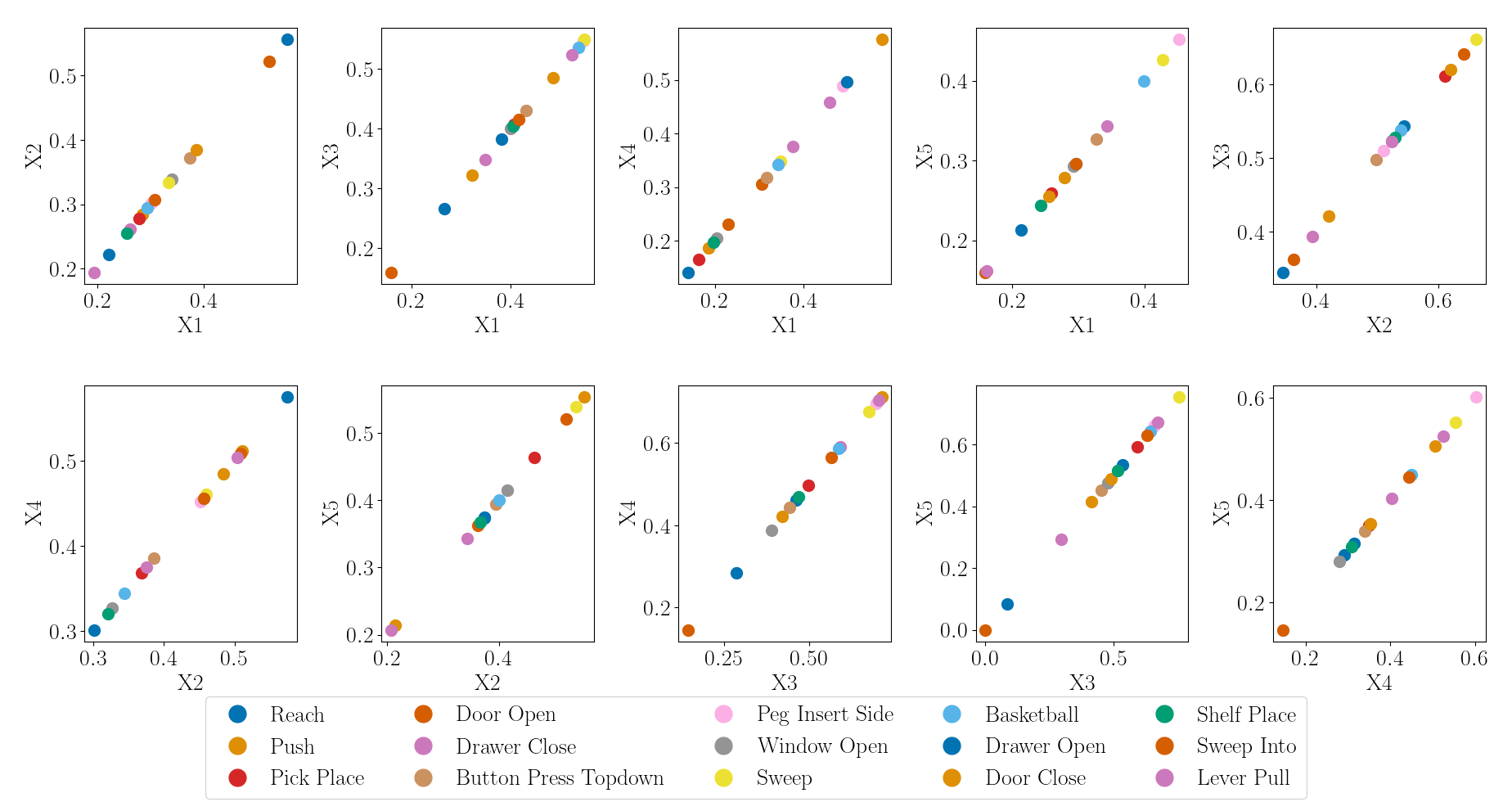}
    \caption{Learned 5-dimensional latent representations associated with different tasks in the MetaWorld ML-10 benchmark. Each pair of two elements of the average task belief vector is plotted against each other.}
    \label{fig:meta-pairwise}
\end{figure}

\begin{figure}[ht!]
    \centering
    \includegraphics[width=\columnwidth]{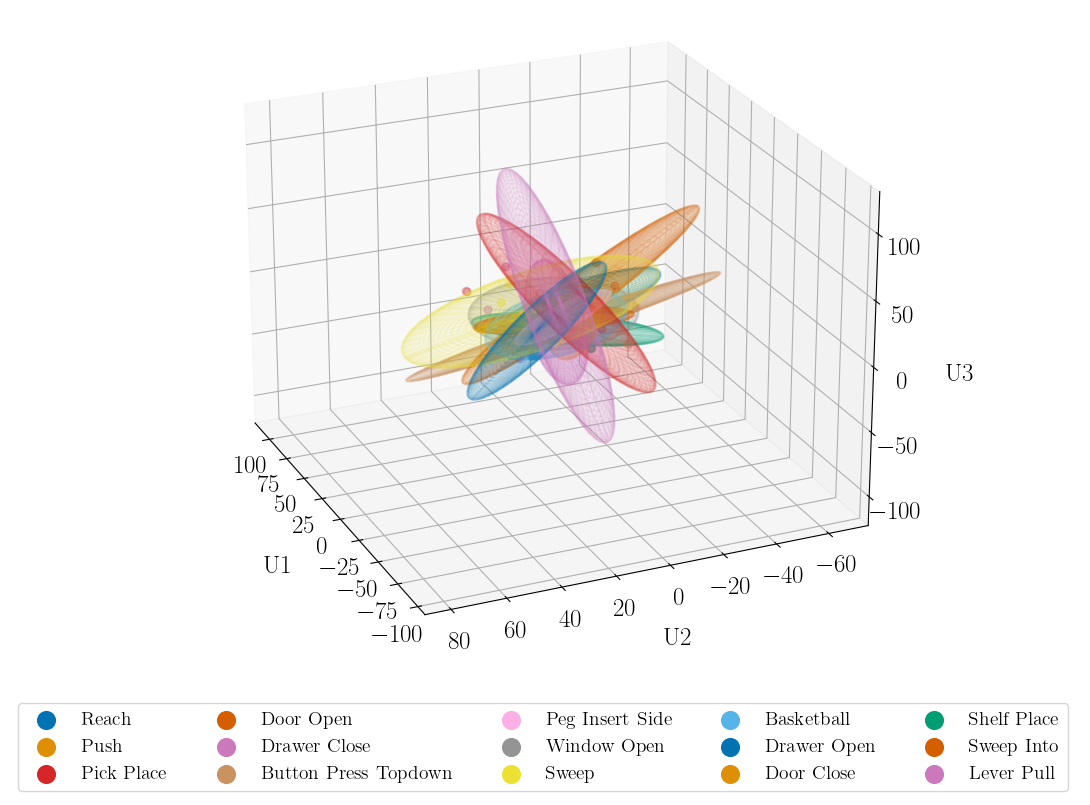}
    \caption{3-D projection of latent representations associated with different tasks in the MetaWorld ML-10 benchmark. The projection is computed using the UMAP method \cite{mcinnes2018umap}. An approximate Gaussian 3-D ellipsoid is illustrated for each task.}
    \label{fig:meta-umap}
\end{figure}

To further analyze the structure of the learned latent space, the pairwise projections of the five-dimensional task belief vectors are plotted. A 2-D plot is generated for each pair of the latent space dimensions, resulting in ten distinct subplots covering all one-vs-one combinations shown in \Cref{fig:meta-pairwise}. Some task environments cluster closely in certain latent dimensions while remaining dispersed in others, suggesting that different latent axes are responsible for encoding distinct aspects of task variation. A 3-D projection of the latent vectors is illustrated in \Cref{fig:meta-umap}. By finding the means and variances of the 3-D UMAP projections computed for the latent representations of each of the 15 tasks \cite{mcinnes2018umap}, the estimated distributions are visualized as ellipsoids, indicating the variability region of each task in the latent space.

\begin{figure}[ht!]
    \centering
    \includegraphics[width=\columnwidth]{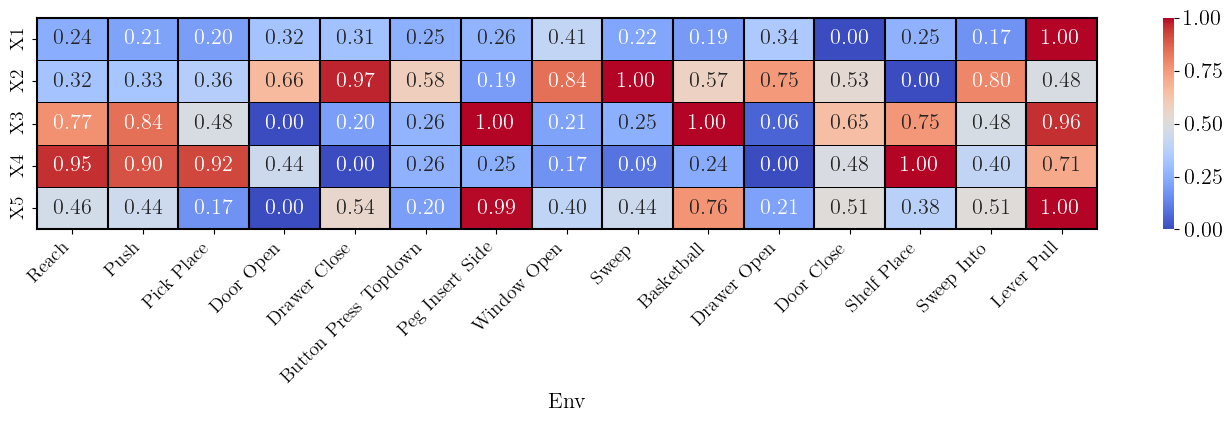}
    \caption{Heatmap of mean task belief vectors inferred by CRAFT over all evaluation episodes in MetaWorld ML-10. Task-specific activation patterns and shared dominant latent dimensions among related tasks demonstrate disentangled representations aligned with semantics without action information.}

    \label{fig:meta-heatmap}
\end{figure}

\Cref{fig:meta-heatmap} shows the average latent vector in all evaluation episodes of all variations of each environment. Some tasks consistently exhibit large values along specific latent dimensions, effectively dominating those axes. For instance, the \textit{lever-pull} task is represented by a very low value along axis one and a large value on axis three, while the \textit{door-open} task shows the highest value on axis one compared to all others. In general, conceptually similar tasks tend to share these dominant dimensions, indicating that the model has learned a structured and interpretable belief space. This organization reflects the transformer encoder-decoder’s ability to disentangle task-specific features. CRAFT was able to extract meaningful representations without access to action information, supporting its role as a reliable inference module in adaptive meta-RL.

\section{Conclusions}
\label{sec:meta_conclusion}

The experiments presented in this paper evaluate CRAFT, the proposed transformer-based belief model for action-free variational task inference in robot manipulation with meta-reinforcement learning. Using a well-maintained benchmark consisting of goal-oriented robotic manipulation tasks, the model is assessed on both adaptation performance and quality of learned task representations.

CRAFT consistently demonstrated strong adaptation performance in training environments, outperforming established baselines such as RL$^2$, MAML, VariBAD, and SDVT by a considerable margin. This means that without action information, Bayes-adaptive sampling of sequences of states and rewards offers sufficient structure to support effective task inference and policy adaptation. While this method maintains competitive performance on generalization and adaptation to test environments, additional virtual training used in SDVT proves beneficial for unseen tasks by training the policy on overlapping belief representations. As a result, SDVT achieves comparable results in hold-out environments, and its virtual training approach can be an effective addition to CRAFT. Nonetheless, the model maintains a significant and consistent advantage in training tasks, reflecting efficient context inference and adaptation in familiar settings.

The analysis of final-episode performance as opposed to meta-episode average highlights the effectiveness of the proposed action-free method in active exploration and leveraging accumulated context. This advantage persists throughout training, especially in intermediate and later stages, suggesting that belief refinement in this model can continue to support improved adaptation. Hence, it was indicated that transformer-based models demonstrate a stronger capacity to extract information from longer sequences compared to memory-based baselines.

Also, based on the ablation study of two different belief models, the inclusion of a state decoder during training, although beneficial in exploration, overregulates the latent representation and harms the adaptation performance. That is, since the context encoder is action-free, the state decoder serves as an unnecessary information bottleneck for its use of actions, which are deemed privileged information in this case. Additionally, it was qualitatively confirmed that the learned latent space shows coherent structure across tasks.

In summary, CRAFT provides a robust and flexible alternative for creating task inference modules in meta-reinforcement learning, especially where action logging is infeasible or undesirable. Its generalization capabilities and action-agnostic architecture make it suitable for real-world deployment in adaptive control systems.

\section{Future Directions}
\label{sec:meta_future}
The method presented here improves on some adaptation metrics and shows a strong quality of task representation. To do so, a major contributing factor to the dynamic definition of an MDP, the actions, was removed from the belief model, and this lack of information was compensated for by the suitable design of a transformer encoder-decoder that captures useful features from a less informative combination of sequences. The assumption under which this redesign was based, i.e., the minimal role of action penalties in reward definition, is present in goal-oriented robotic tasks. However, the same approach is limited in application by the foundational concepts of what is deemed \textit{success} in other domains. Additionally, transformers, even a single-layer architecture as used for this belief model, are notorious for introducing computational overhead, particularly during inference.

Considering these limitations, future research is now possible based on the flexible design proposed and validated here. The abundance of action-free robot manipulation trajectories has energized pre-training and semi-supervised learning in offline reinforcement learning from image observations \cite{seo2022reinforcement, zheng2023semi}. This shows the importance of using numerous datasets of recorded demonstrations to facilitate generalizability in robotic manipulation. The proposed action-free belief model, trained independently from the RL agent, holds promise as a tool for providing RL agents with auxiliary task representations to support adaptation to new tasks.

\begin{figure}[ht!]
    \centering
    \includegraphics[width=\columnwidth]{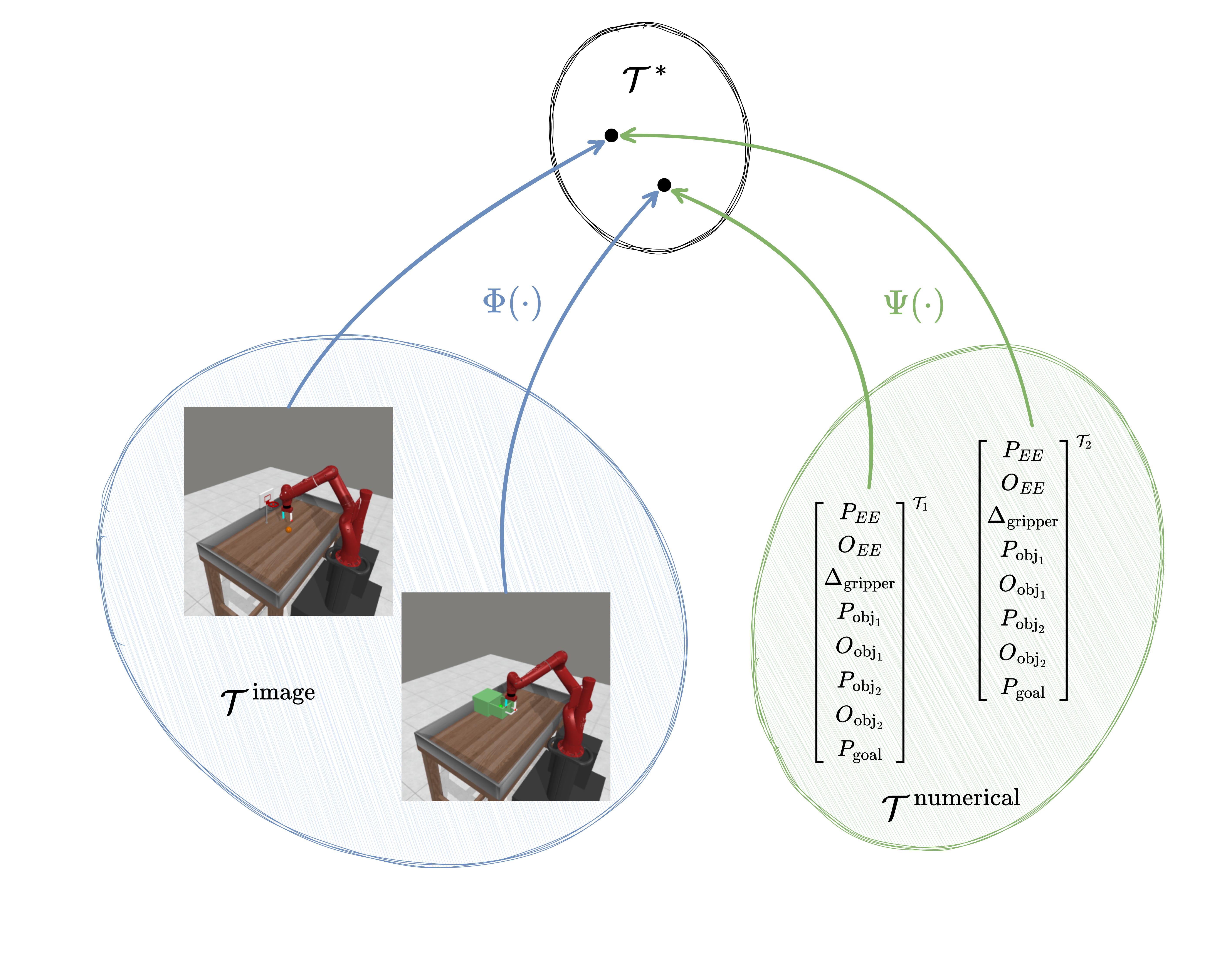}
    \caption{A conceptual shared task latent representation for MDPs with different state observation modalities}
    \label{fig:meta-mapping}
\end{figure}

Considering the mentioned line of research, two possibilities may emerge from using the inherent flexibility of CRAFT in this paper:
\begin{enumerate}
    \item Pre-training the belief model with action-free video recordings of various robot manipulation tasks, which requires more training investment because of the transformer architecture, and then online adaptation to a new task in a similar environment.
    \item Cross-modal adaptation to environments with different state observation modes.
\end{enumerate}

\begin{figure}[ht!]
    \centering
    \includegraphics[width=\columnwidth]{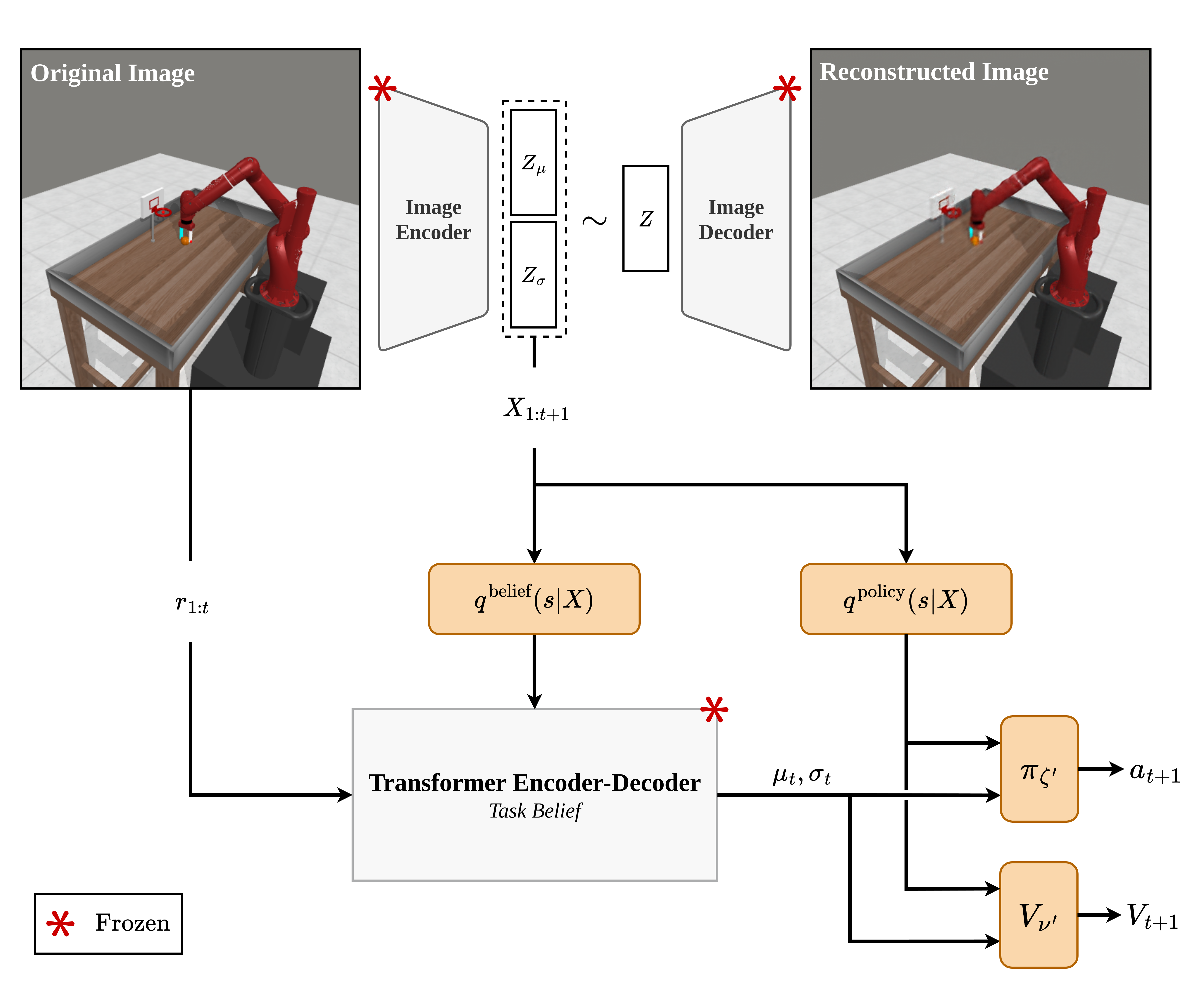}
    \caption{Outline of a possible research on cross-modal adaptation scenario}
    \label{fig:meta-num2img}
\end{figure}

The former possibility is quite straightforward in terms of implementation. However, the latter needs further explanation. As an example, consider a robotic environment expressed by two different MDPs $\mathcal{M}_1$ and $\mathcal{M}_2$ per \Cref{eq:meta-mdp}. These two MDPs share the same action space ($\mathcal{A}$), dynamic functions ($\mathcal{P}_S,  \mathcal{P}_R$), discount factor ($\gamma$), and episode length ($H$). However, the two MDPs differ in the state space domains $\mathcal{S}_1$ and $\mathcal{S}_2$ and optimal policies $\pi^*_1$ and $\pi^*_2$, respectively. For instance, $\mathcal{M}_1$ can receive state observations in the numerical space from sensor data, versus $\mathcal{M}_2$ receiving workspace images as observations. Assuming the essential Markov property holds for both MDPs, it is possible to infer a shared latent domain where $\Psi(\mathcal{S}_1) \cong \Phi(\mathcal{S}_2)$, as conceptualized in \Cref{fig:meta-mapping}, such that $\Psi(\mathcal{M}_1) \simeq \Phi(\mathcal{M}_2)$. These transformed MDPs can be governed with a shared optimal policy $\pi^* \simeq \pi^*(\Psi(\cdot)) \simeq \pi^*(\Phi(\cdot))$.

If learning the mappings $\Psi(\cdot)$ and $\Phi(\cdot)$, and their inverse functions $\Psi^{-1}(\cdot)$ and $\Phi^{-1}(\cdot)$, is tractable, it is possible to train an action-agnostic, and hence agent-agnostic,  foundational transformer encoder-decoder from offline data in the source domain and fine-tune it with additional adapters to enable adaptation in the target domain. This means the proposed belief model can be trained on an MDP with numerical state observations. By learning observation adapters $q^\text{belief}$ and $q^\text{policy}$, and fine-tuning the RL agent consisting of value function $V_\nu$ and $\pi_\zeta$, the context-adaptive meta-RL pipeline can acclimate to visual state observations online as outlined in \Cref{fig:meta-num2img}.

To reiterate, the inherently less structured design of CRAFT enables its flexible application to multi-task robotic manipulation via offline pre-training or cross-modal adaptation. Nonetheless, action-free task inference with a transformer encoder-decoder enables flexible adaptation in robotic manipulation environments where actions may be unavailable, expensive to track, or noisy.

\bibliographystyle{unsrtnat}
\bibliography{./references}

\end{document}